\documentclass{article} %
\usepackage{iclr2027_conference,times}
\iclrfinalcopy

\usepackage{amsmath,amsfonts,bm}

\def\eqref#1{equation~\ref{#1}}

\def\1{\bm{1}}

\DeclareMathAlphabet{\mathsfit}{\encodingdefault}{\sfdefault}{m}{sl}
\SetMathAlphabet{\mathsfit}{bold}{\encodingdefault}{\sfdefault}{bx}{n}

\usepackage{hyperref}
\usepackage{url}
\usepackage{graphicx}
\usepackage{twemojis}
\DeclareRobustCommand{\emojiSup}[1]{%
  \begingroup
  \setbox0=\hbox{\twemoji[height=\fontcharht\font`X]{#1}}%
  \smash{\textsuperscript{\copy0}}%
  \endgroup
}
\DeclareRobustCommand{\realagent}{\emojiSup{laptop}}
\DeclareRobustCommand{\vOSS}{\emojiSup{unlock}}

\usepackage{listings}
\usepackage{enumitem}
\usepackage{xcolor}  %
\usepackage{colortbl}  %
\usepackage{xurl} %
\usepackage{algorithmic} %
\usepackage{booktabs}
\usepackage{siunitx}
\usepackage{array}
\usepackage[export]{adjustbox}
\newcommand{\normallink}[2]{%
  \href{#1}{\small{\nolinkurl{#2}}}%
}

\usepackage[table]{xcolor}

\definecolor{deltagain}{HTML}{1A7F37}
\definecolor{deltadrop}{HTML}{C0392B}
\definecolor{oursbg}{HTML}{F3F6FB}

\lstdefinestyle{pythonstyle}{
    language=Python,
    basicstyle=\ttfamily\scriptsize,
    keywordstyle=\color{blue}\bfseries,
    stringstyle=\color{red},
    commentstyle=\color{gray}\itshape,
    numbers=left,
    numberstyle=\tiny\color{gray},
    numbersep=5pt,
    breaklines=true,
    breakatwhitespace=false,
    showstringspaces=false,
    frame=single,
    framesep=2pt,
    xleftmargin=15pt,
    framexleftmargin=15pt,
    tabsize=4,
    columns=flexible,
    keepspaces=true,
}

\usepackage{microtype}

\usepackage{xspace}
\usepackage{multirow}
\usepackage{diagbox}
\usepackage{balance}
\usepackage{lipsum}
\usepackage{wrapfig}
\usepackage{tabularx}
\usepackage{subcaption}

\SetTracking[no ligatures = f]{encoding = * , shape = sc}{-80}

\definecolor{benchgray}{HTML}{EEF1F6}

\input{paragraph_refs.tex}

\newcommand{\ours}{{\small\sffamily WildAWM}\xspace}

\newcommand{\passtok}{\text{pass\^{}k}\xspace}

\newcommand{\passatk}[1]{\text{pass@#1}\xspace}

\title{Agent World Model in the Wild: \\ Computer-Use Agent Environment Synthesis}

\hypersetup{pdfauthor={Zhaoyang Wang, Qianhui Wu, Haorui Wang, Simon Yu, Baolin Peng, Canwen Xu, Hao Cheng, Wenlin Yao, Boyi Liu, Yite Wang, Jianfeng Gao, Huaxiu Yao}}
\hypersetup{pdftitle={Agent World Model in the Wild: Computer-Use Agent Environment Synthesis}}

\author{%
\scalebox{0.96}{
Zhaoyang Wang$^{1,\ast}$,
Qianhui Wu$^{2,\dagger}$,
Haorui Wang$^{3,\ast}$, Simon Yu$^{4,\ast}$, Baolin Peng$^{2}$}\\
\scalebox{0.96}{
\bf Canwen Xu$^{5}$,
Hao Cheng$^{2}$,
Wenlin Yao$^{2}$,
Boyi Liu$^{5}$,
Yite Wang$^{5}$,
Jianfeng Gao$^{2}$,
Huaxiu Yao$^{1}$}\\[0.5em]
$^{1}$UNC-Chapel Hill\quad$^{2}$Microsoft Research\quad$^{3}$Georgia Tech\\
$^{4}$Northeastern University\quad$^{5}$Independent Researcher\\[0.3em]
{\tt\small zhaoyang@cs.unc.edu~~~qianhuiwu@microsoft.com}\\[3pt]
\wildawmcentred{\small\strut
\wildawmlink{https://msr-orchard.github.io/WildAWM/}{\faGlobe}{Project Page}\qquad
\wildawmlink{https://huggingface.co/collections/ChilleD/agent-world-model-in-the-wild}{\hflogo}{Data \& Models}\qquad
\wildawmlink{https://hub.harborframework.com/datasets/wildawm}{\harborlogo}{Harbor}\qquad
\wildawmlink{https://github.com/MSR-Orchard/WildAWM}{\faGithub}{Code}}}

\makeatletter
\gdef\@thanks{%
  \protect\footnotetext[1]{Work done during internships at Microsoft Research.\quad $^{\dagger}$Corresponding author.} %
}
\makeatother

\definecolor{unccloud}{HTML}{EDF5FB} %
\newcommand{\wildawmlogo}[1]{\raisebox{\dimexpr 8pt-0.5\height\relax}{#1}}
\fancypagestyle{wildawmfirstpage}{%
  \fancyhf{}%
  \fancyhead[R]{\smash{%
    \wildawmlogo{\includegraphics[height=15.5pt]{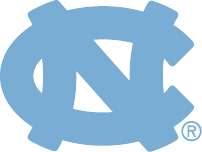}}\hspace{10pt}%
    \wildawmlogo{\includegraphics[height=16pt]{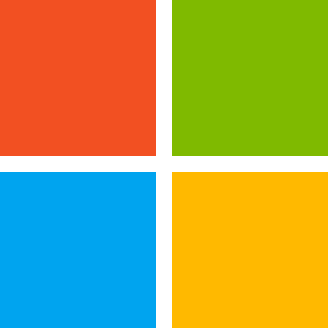}}\hspace{10pt}%
    \wildawmlogo{\includegraphics[height=14.5pt]{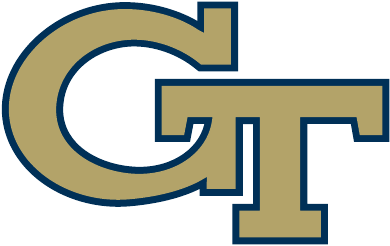}}\hspace{10pt}%
    \wildawmlogo{\includegraphics[height=15pt]{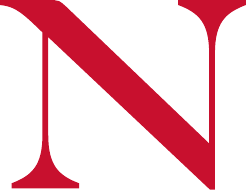}}}}%
  \fancyfoot[C]{\thepage}%
  \renewcommand{\headrulewidth}{0pt}%
}
\usepackage{fontawesome5} %
\definecolor{uncbolincreek}{HTML}{2C5080} %
\newcommand{\hflogo}{\raisebox{-0.15em}{\includegraphics[height=1.05em]{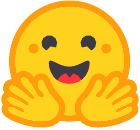}}}
\newcommand{\harborlogo}{\raisebox{-0.15em}{\includegraphics[height=1em]{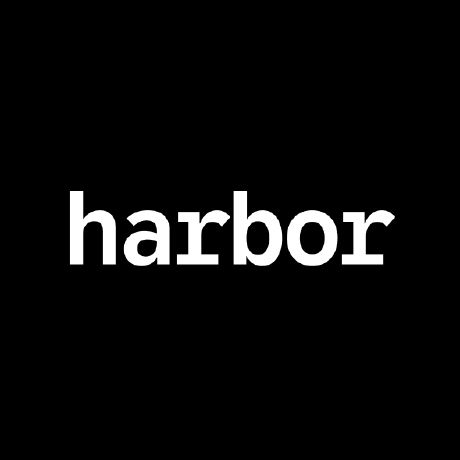}}}
\newcommand{\wildawmlink}[3]{{\hypersetup{pdfborder={0 0 0}}%
  \href{#1}{\smash{#2}\ \textcolor{uncbolincreek}{#3}}}}
\newcommand{\wildawmcentred}[1]{\hspace*{-\tabcolsep}\rlap{\makebox[\textwidth][c]{#1}}}
\usepackage{etoolbox}
\makeatletter
\patchcmd{\@maketitle}{\hsize\textwidth}
  {\hsize\textwidth\vskip-\dimexpr\baselineskip+3pt\relax\vskip-\parskip}{}
  {\PackageWarningNoLine{wildawm}{Could not raise the title block}}
\makeatother
\usepackage{tikz}
\usetikzlibrary{tikzmark,calc}
\newcommand{\wildawmabstractpanel}{%
  \tikz[remember picture, overlay]{%
    \iftikzmark{wildawm-abs-top}{\iftikzmark{wildawm-abs-bot}{%
      \coordinate (t) at (pic cs:wildawm-abs-top);%
      \coordinate (b) at (pic cs:wildawm-abs-bot);%
      \fill[unccloud, rounded corners=2mm]
        ($(t) + (0pt, 14pt)$) rectangle ($(t |- b) + (\linewidth, -8pt)$);%
    }{}}{}}}
\renewenvironment{abstract}
  {\vskip.075in
   \hbox to\linewidth{\tikzmark{wildawm-abs-top}\wildawmabstractpanel\hss{\large\sc Abstract}\hss}%
   \vspace{0.5ex}\begin{quote}}
  {\ifhmode\unskip\tikzmark{wildawm-abs-bot}\fi\par\end{quote}\vskip 1ex}

\newcommand{\ie}{\textit{i.e.}}
\newcommand{\eg}{\textit{e.g.}}
\begin{document}

\maketitle
\thispagestyle{wildawmfirstpage}

\begin{abstract}
Agent learning is shifting from learning from human data toward learning through interaction with the world.
However, scaling such learning directly in real-world computer systems is challenging and can introduce substantial security risks.
We therefore propose \textbf{A}gent \textbf{W}orld \textbf{M}odel in the \textbf{Wild} (\ours), a scalable pipeline for synthesizing controllable environments and tasks for training general computer-use agents (CUAs).
Whereas existing approaches typically specialize in a single agent interaction interface, we construct executable environments that jointly support graphical user interfaces, model context protocol services, and command-line interfaces, allowing agents to flexibly combine visual interaction, tool use, and code execution.
\ours is entirely powered by open-source models, demonstrating that high-quality computer-use environments can be generated without proprietary frontier models. 
Beyond environments, we address three recurring challenges in task synthesis: \textit{ensuring solvability, maintaining sufficient difficulty, and producing reliable verifiers}, by introducing execution-based quality gates that combine heterogeneous agent rollouts, cross-judge consistency checks, and counterfactual probes of judgment robustness.
The resulting pipeline synthesizes 341 long-horizon environment-task pairs that can require over 100 agent steps to solve and remain challenging even for frontier models, with GPT-6-Astra solving only 60\% of them.
We train two Qwen models of different sizes exclusively in these synthetic worlds. Both models achieve substantial gains across 8 out-of-distribution benchmarks spanning web, claw, and computer-use agents, demonstrating that learning in \ours's synthesized worlds generalizes broadly to unseen computer-use tasks.
The code is publicly available at \href{https://github.com/MSR-Orchard/WildAWM}{https://github.com/MSR-Orchard/WildAWM}.
\end{abstract}

\section{Introduction}
\label{sec:intro}

Large language model (LLM) powered agents are shifting from learning from humans to learning from the world~\citep{silver2025welcome}. Early agents learned from human data, which are expensive to collect and limited by demonstrator's ability. Recent agents instead gain much of their capability through reinforcement learning (RL) in executable environments~\citep{kimik3,glm5,deepseekv4}, making environment supply a key bottleneck.
However, training directly in the real world is costly and risky: states are hard to reset, online services change over time, and exploratory actions may have unintended consequences.\footnote{For example, see OpenAI's recent incident report on real-world RL at \href{https://openai.com/index/hugging-face-incident-and-the-road-ahead/}{this url}.}
Manually built environments restore control and verifiability~\citep{taubench,zhou2024webarena,xie2024osworld}, but remain expensive and hard to scale.
Synthetic environments and tasks offer a path to controllable, large-scale online RL.

Computer-Use Agents (CUAs) represent a broad setting for environment synthesis, as they operate across applications, websites, and local files on a personal computer.
A single task may be completed through graphical user interface (GUI) interaction, command-line interface (CLI) operations, or calls to model context protocol (MCP) servers, making interface selection and coordination an important capability.
Existing synthesis pipelines often target individual interaction modes, including browser use~\citep{zhou2026wainf,wu2026autowebworld}, tool calling~\citep{du2026generalizable,wang2026agentworldmodelinfinity}, and terminal workflows~\citep{li2026recursive,zhu2026termigen}. 
Prior CUA work, however, often focuses primarily on GUI interaction~\citep{aggarwal2026gym,wang2026cuagymscalingverifiabletraining}. 
In this paper, we adopt a broader view of computer use spanning common interfaces on a personal computer, since real computer-use tasks are defined by what can be accomplished rather than by a single interaction modality.

Synthesizing environments and tasks for this broader CUA setting introduces challenges beyond prior synthesis work. Existing approaches often synthesize visually simplified applications and build each application as an isolated environment.
In contrast, realistic computer use spans a connected digital workspace in which websites, applications, MCP services, and local files coexist within an executable personal computer environment.
Meanwhile, synthesized tasks should be \textit{solvable} under the synthesized  environment state, sufficiently \textit{challenging} to provide useful learning signals, and paired with \textit{reliable} verifiers to distinguish successful completion from incorrect outcomes, all without relying on simplistic assumptions such as ``a frontier model is necessarily correct.''

To address these challenges, we propose \ours, a scalable synthesis pipeline for general CUAs. Starting from user personas and daily activities, \ours constructs executable personal-computer environments with websites, MCP services, and local files, then synthesizes cross-agent interface (XAI) tasks grounded in these worlds.
We introduce execution-based multi-agent quality gates, where heterogeneous agents from different models and harnesses execute each task, providing evidence of solvability, difficulty, and judge consistency.
To further test verifier robustness, we exploit LLM sycophancy~\citep{sharma2024towards} through counterfactual probes with deliberately incorrect outcome feedback.
The resulting environments, tasks, and verifiers provide executable worlds and rewards for agentic RL.
Notably, \ours uses open-source models throughout synthesis, whereas existing works often rely on proprietary models~\citep{aggarwal2026gym,zhou2026wainf,wang2026cuagymscalingverifiabletraining}. This improves reproducibility, transparency, and accessibility, while showing that open-source models alone can synthesize high-quality environments and challenging tasks for general CUA training.

Using \ours, we synthesize 341 environment--task pairs, each requiring at least three interface types, with long-horizon workflows that may require over 100 agent steps, and even frontier model GPT-6-Astra~\citep{gpt6} solves only about 60\%.
We further perform online RL on two Qwen models, Qwen3.6-35B-A3B~\citep{qwen36_35b_a3b} and Qwen3.5-4B~\citep{qwen35blog}, using OpenCode CLI harness~\citep{opencode2025} equipped with GUI and MCP tools, and find that training exclusively in these synthetic worlds yields substantial gains across all eight out-of-distribution (OOD) benchmarks spanning web, claw, and computer-use agents.
Our contributions are threefold: 
\begin{itemize}[leftmargin=0.5cm, nosep]
  \item We propose a scalable, fully open-source pipeline \ours for synthesizing CUA environments spanning GUI, CLI, and MCP interfaces.
  \item We propose execution-based gates for synthesizing solvable, challenging tasks with reliable verifiers, together with the release of 341 environment--task pairs.
  \item  Experiment results demonstrate that training exclusively in worlds synthesized by open-source models can substantially improve the generalization performance of computer-use agents. 
\end{itemize}

\begin{table}[t]
  \centering
  \caption{Comparison of related methods.  Scope reports the synthesized task setting. PC denotes personal computer. Modality denotes primary target-agent observations. Synthesis Source darkens with more real world dependence. $^*$ WebHarbor's websites provide \ours's seed GUI substrate and image content. \vOSS denotes open-source models described in \parref{par:method:rollouts-judgments}, while other methods all use proprietary models. \realagent denotes that the target agent is often equipped with GUI and MCP tools, thus being able to interact with all interfaces on a personal computer.
  }
  \label{tab:related_work_comparison}
  \small
  \setlength{\tabcolsep}{3pt}
  \definecolor{rwgreen}{HTML}{1A7F37}
  \definecolor{rwred}{HTML}{C0392B}
  \definecolor{srcscratch}{HTML}{F7FBFF}
  \definecolor{srcseed}{HTML}{DEEBF7}
  \definecolor{srcreference}{HTML}{C6DBEF}
  \definecolor{srcsoftware}{HTML}{9ECAE1}
  \newcommand{\cmark}{\textcolor{rwgreen}{\ensuremath{\checkmark}}}
  \newcommand{\xmark}{\textcolor{rwred}{\ensuremath{\times}}}
  \newcommand{\vlogo}[1]{\textsuperscript{\raisebox{-0.15ex}{#1}}}
  \newcommand{\vOpenAI}{\vlogo{\includegraphics[height=2.6ex]{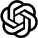}}}
  \newcommand{\vAnthropic}{\vlogo{\includegraphics[height=2.6ex]{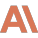}}}
  \newcommand{\vGemini}{\vlogo{\includegraphics[height=2.6ex]{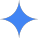}}}

  \begin{adjustbox}{max width=\linewidth}
    \begin{tabular}{@{}lcccccccc@{}}
    \toprule
    \multirow{2}{*}{\textbf{Method}} & \multirow{2}{*}{\textbf{Scope}} & \multirow{2}{*}{\textbf{Modality}} & \multicolumn{3}{c}{\textbf{Interfaces}} & \multirow{2}{*}{\shortstack{\textbf{Synthesis}\\\textbf{Source}}} & \multirow{2}{*}{\shortstack{\textbf{Synthesis}\\\textbf{Models}}} & \multirow{2}{*}{\textbf{Target Agent}} \\
    \cmidrule(lr){4-6}
    & & & GUI & MCP & CLI & & & \\
    \midrule
    AWM~\citep{wang2026agentworldmodelinfinity} & Single App & Text & \xmark & \cmark & \xmark & \cellcolor{srcscratch}From-Scratch & GPT-5\vOpenAI & ReAct-style Tool-use Agent \\
    WebArena-Infinity~\citep{zhou2026wainf} & Single App & Text & \cmark & \xmark & \xmark & \cellcolor{srcreference}Reference-Grounded & Opus 4.6\vAnthropic & ReAct-style Web Agent \\
    Gym-Anything~\citep{aggarwal2026gym} & Single App & Text+Image & \cmark & \xmark & \xmark & \cellcolor{srcsoftware}Existing Software & Opus 4.5/4.6\vAnthropic & ReAct-style GUI Agent \\
    TMax~\citep{ivison2026tmaxsimplerecipeterminal} & Multi App & Text & \xmark & \xmark & \cmark & \cellcolor{srcseed}Seed-Grounded & Gemini-3-Pro\vGemini & mini-SWE-agent \\
    Synth. Computers~\citep{ge2026syntheticcomputersscalelonghorizon} & PC & Text & \xmark & \xmark & \cmark & \cellcolor{srcscratch}From-Scratch & Opus 4.6\vAnthropic & Claude Code\realagent \\
    CUA-Gym~\citep{wang2026cuagymscalingverifiabletraining} & Multi App & Text & \cmark & \xmark & \xmark & \cellcolor{srcreference}Reference-Grounded & Sonnet 4.6\vAnthropic & ReAct-style GUI Agent \\
    CUA-Universe~\citep{shi2026cuauniverse} & Single App & Text+Image & \cmark & \xmark & \cmark & \cellcolor{srcsoftware}Existing Software & GPT-5.6\vOpenAI & Path-Steer GUI+CLI Agent \\
    \midrule
    \textbf{\ours} & PC & Text+Image$^*$ & \cmark & \cmark & \cmark & \cellcolor{srcseed}From-Scratch + Seed-Grounded$^*$ & OpenSource\vOSS & OpenCode\realagent \\
    \bottomrule
  \end{tabular}
  \end{adjustbox}
\end{table}

\section{Related Work}
\label{sec:related_work}

\myparagraph[par:related-work:computer-use-agents]{Computer-Use Agents.}
Computer-use agents interact with applications through GUIs~\citep{cheng2024seeclick,xu2024aguvis,wu2025guiactor} and programmatic representations such as HTML and accessibility trees~\citep{zhou2024webarena,he2024webvoyager}. GUI agents have been studied for web navigation~\citep{he2024webvoyager,zheng2024seeact,yang2026openwebrldemystifyingonlinemultiturn}, desktop automation~\citep{agashe2025agent,zhang2025ufo,zhang2025ufo2,wang2026opencua}, and multi-application workflows~\citep{tan2024cradle,qin2025uitars,openaicua,yang2026guilibra}. 
Recent work expands this action space with CLI~\citep{qwenuiagent}, coding~\citep{coact}, and MCP tools~\citep{jia2025osworldmcpbenchmarkingmcptool}, while harness-based agents expose heterogeneous tools through coding and claw-agent frameworks~\citep{yu2026openforgerltrainharnessnativeagents,openclaw,claudecode,opencode2025}. 
These trends suggest a broader view of computer use in which the agent is not bound to a particular interaction modality. In this work, we adopt this view and consider general CUAs that can jointly access GUI, CLI, and MCP interfaces as needed to complete any tasks on a personal computer.

\myparagraph[par:related-work:environment-synthesis]{Environment Synthesis.}
The growing demand for interactive agent training has motivated automatic construction of executable environments.
Recent work synthesizes executable environments for web interaction~\citep{zhou2026wainf,wu2026autowebworld}, tool use~\citep{wang2026agentworldmodelinfinity,du2026generalizable,dong2026agent}, terminal interaction~\citep{zhu2026termigen,ivison2026tmaxsimplerecipeterminal,li2026recursive}, and software engineering~\citep{jain2025r2egymproceduralenvironmentshybrid,yang2025swesmithscalingdatasoftware,chen2026sweuniversescalerealworldverifiable}. 
For CUAs, GUI-GENESIS reconstructs lightweight web applications~\citep{guigenesis}, WebArena-Infinity and CUA-Gym synthesize web environments and verifiable tasks~\citep{zhou2026wainf,wang2026cuagymscalingverifiabletraining}, Gym-Anything converts existing software into agent environments~\citep{aggarwal2026gym} rather than building the environments, and CUA-Universe augments desktop applications with CLI access~\citep{shi2026cuauniverse}. Other work synthesizes user-specific computers with productivity artifacts~\citep{ge2026syntheticcomputersscalelonghorizon}. 
As compared in Table~\ref{tab:related_work_comparison}, \ours constructs a unified personal-computer world with joint GUI, MCP, and CLI access, combining WebHarbor's visual web substrate~\citep{webharbor2026} with synthesized services, files, and application data. Notably, \ours uses open-source models instead of proprietary frontier models, which can improve transparency and accessibility for the community.

\myparagraph[par:related-work:task-synthesis]{Task Synthesis.}
Task generation has progressed from standalone instruction synthesis~\citep{wang2023selfinstructaligninglanguagemodels,xu2025wizardlmempoweringlargepretrained,alpaca} to tasks grounded in executable environments~\citep{sun2024osgenesis,pahuja2025explorerscalingexplorationdrivenweb,wang2025adaptingwebagentssynthetic}. Prior methods derive tasks from observed interactions~\citep{sun2024osgenesis,su2025learnbyinteractdatacentricframeworkselfadaptive}, environment exploration~\citep{pahuja2025explorerscalingexplorationdrivenweb,wang2025adaptingwebagentssynthetic}, or programmatic generation~\citep{yang2025swesmithscalingdatasoftware}.
More recent methods couple task synthesis with executable verification: WebArena-Infinity synthesizes difficulty-graded browser tasks with programmatic verifiers~\citep{zhou2026wainf}, Agent World Model (AWM)~\citep{wang2026agentworldmodelinfinity} exploits executable environment states to construct reliable rewards, CUA-Gym~\citep{wang2026cuagymscalingverifiabletraining} co-generates task, environment state, and reward as verified triples.
Beyond these approaches, \ours introduces a quality-control pipeline using heterogeneous agent rollouts, cross-judge consistency, and counterfactual probes to validate task solvability and difficulty while stress-testing the reliability of synthesized verifiers, all with open-source models.

\begin{figure}
    \centering
    \includegraphics[width=1.0\linewidth,height=6.6cm]{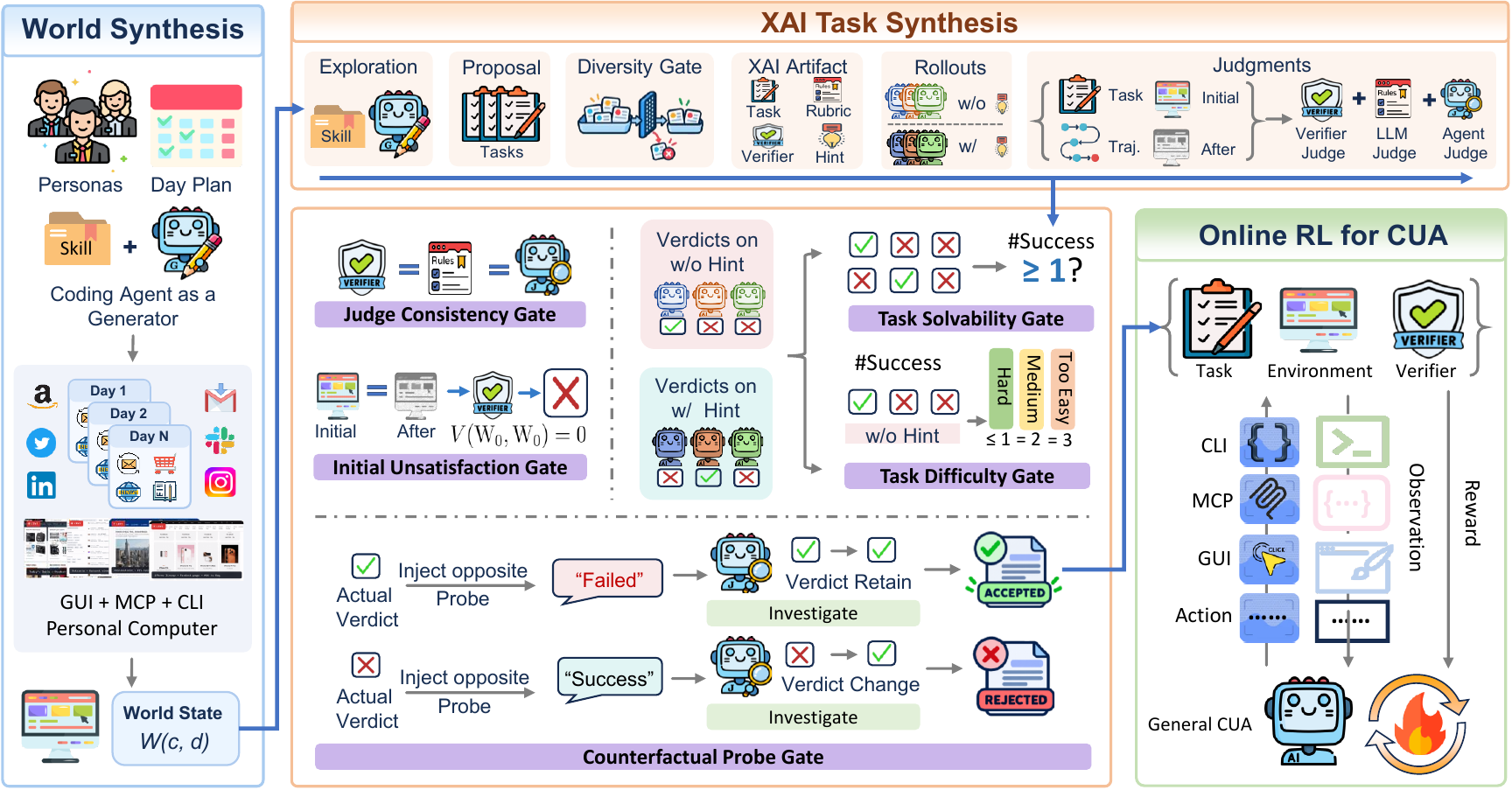}
    \caption{Overview of \ours. The synthesis pipeline includes two stages: (1) \textit{World Synthesis}, where we use the coding agent to construct persona-driven, multi-day world states for computer environments that expose GUI, MCP, and CLI interfaces. (2) \textit{XAI Task Synthesis}, where we generate cross-interface tasks through 5 designed quality gates for task solvability, difficulty, and verifier reliability. The synthesized $\small \{\texttt{task}, \texttt{environment}, \texttt{verifier}\}$ triplet enables training general CUA.}
    \label{fig:main}
    \vspace{-0.3cm}
\end{figure}

\section{Agent World Model in the Wild}
\label{sec:method}
As shown in Figure~\ref{fig:main}, \ours uses open-source models to synthesize executable environments and tasks for training general CUAs. Each synthesized instance contains a personal-computer environment initialized from a world state $W$, a task $\tau$, and a verifier $V$ that evaluates task completion.

\subsection{World Synthesis}
\label{sec:method:world}
We define a world as the accumulated state of a personal computer which comprises web applications, MCP-accessible services, and local files that record a person's activities.
To expand the range of multimodal GUI interaction, we integrate 15 mirror websites from WebHarbor~\citep{webharbor2026} which provides website mirrors with rich visual content.
We first synthesize a persona and expand its ongoing activities into sequential day plans. Then, a coding agent materializes these plans day by day, with each day extending the previous state.
This procedure constructs applications and services, and produces diverse user histories, application data, and files.
We denote the cumulative state of persona $c$ after day $d$ as $W(c,d)$, which initializes and resets the environment state.

\myparagraph[par:method:persona-day-planning]{Persona and Day Planning.}
We first prompt the generator, a coding agent using GLM-5.2~\citep{glm5} within OpenCode~\citep{opencode2025}, to synthesize personas from scratch.
Each persona $c$ describes an occupation, named contacts, frequently used websites, and ongoing activities spanning multiple days.
To promote diversity, we compare each candidate with accepted personas in terms of life domain and multi-day storylines, then regenerate repeated candidates with feedback on identified similarities.
For each accepted persona, the generator further develops these storylines into a sequence of day plans.
Each plan specifies activities, the people and digital artifacts involved, the required applications and services, and the corresponding databases and files.
Planning proceed one day at a time: the generator would inspect previous day's state $W(c,d-1)$ to maintain continuity with ongoing activities before planning day $d$.
Then, the subsequent materialization stage would reuse existing application and service implementations and synthesize any artifacts that are missing.

\begin{wraptable}{r}{0.48\textwidth}
  \centering
  \setlength{\belowcaptionskip}{4pt}
  \caption{World synthesis statistics. ``touched'' means the number of services or databases that are modified during the day of synthesis.}
  \label{tab:world_syn_basics}
  \scriptsize
  \setlength{\tabcolsep}{1.25pt}
  \resizebox{\linewidth}{!}{%
  \begin{tabular}{@{}lr@{\hspace{4pt}}lr@{}}
    \toprule
    \multicolumn{2}{c}{\itshape Scale and diversity} & \multicolumn{2}{c}{\itshape Applications} \\
    \cmidrule(lr){1-2}\cmidrule(lr){3-4}
    Personas & 36 & Websites/MCPs & 15/8 \\
    World states & 640 & Reusable skills & 83 \\
    States per persona & 17.8 & Sites/MCPs per persona & 5.25/5.44 \\
    \midrule
    \multicolumn{2}{c}{\itshape Daily mean} & \multicolumn{2}{c}{\itshape Final-day mean} \\
    \cmidrule(lr){1-2}\cmidrule(lr){3-4}
    Sites/MCPs touched & 3.97/5.31 & DB records & 171.6K \\
    DB tables/files touched & 15.38/4.97 & \# Local files & 50.1 \\
    World size growth & 30.9\,KiB & World state size & 59.3\,MiB \\
    \bottomrule
  \end{tabular}%
  }
\end{wraptable}

\myparagraph[par:method:skill-enhanced-generation]{Skill-Enhanced Generation.}
Given the day plan, the generator first matches its activities and required services against a growing library of reusable event skills and application implementations.
Each event skill specifies the interfaces involved, the state changes to produce, and the consistency constraints across the resulting records and files.
For example, a skill for a student enrolling in a Coursera course and completing assignments can encode the corresponding workflow and the required database and file updates.
The generator synthesizes a new skill if a plan introduces an activity not being covered by any existing skill.
Similarly, when a required MCP service is unavailable, it implements the corresponding database schema and backend, MCP servers.
These skills and service implementations will be added to the shared library for reuse in subsequent days and personas.
The generator starts from a clean state on day 0 or a copy of $W(c,d-1)$ thereafter.
It inspects the actual schemas and existing state, then writes a deterministic, replayable program that updates databases for MCP services and websites, and local files to produce $W(c,d)$.
We retain such state only if two independent replays produce the same state, and all services and websites remain healthy.
Further, we launch another coding agent to verify the state is consistent with the day plan, and provide feedback for necessary revisions.
Our synthetic world artifacts are summarized in Table~\ref{tab:world_syn_basics}, with full pipeline details in Appendix~\ref{appendix:world_synth}.

\subsection{XAI Task Synthesis}
\label{sec:method:task}
Beyond environment synthesis, task generation faces three key challenges for agentic learning: \textit{ensuring solvability}, \textit{maintaining sufficient difficulty}, and \textit{generating reliable verifiers}. We address them with an execution-based multi-agent validation loop, where heterogeneous open-source agents attempt each task and five quality gates use the resulting evidence for acceptance or revision.

\myparagraph[par:method:task-generation]{Task Generation.} Given an initial world state $W_0=W(c,d)$, the skill-enhanced generator explores the live environment and proposes cross-agent interface (XAI) tasks grounded in the available application data, service records, and local files. Each task requires the agent to use all three interfaces, i.e., GUI, MCP, and CLI. A diversity gate selects proposals that are least similar to tasks already accepted or being developed by concurrent workers. For each selected proposal, the generator produces a task specification $\tau$, an executable verifier $V$, a rubric $R$, and a hint $h$, and a document that decomposes $\tau$ into an ordered sequence of sub-tasks.
The verifier takes the initial world state $W_0$, the final world state $W$, and the agent's final answer $a$ as inputs. It evaluates a predefined checklist of $N$ binary criteria, where each criterion is given by $v_i(W_0,W,a)\in\{0,1\}$. We define task success as satisfying all criteria: $V(W_0,W,a)=\prod_{i=1}^{N}v_i(W_0,W,a)$.
Individual criterion scores can provide granular feedback and value estimates for long-horizon RL training.

\begin{wrapfigure}{r}{0.45\textwidth}
  \vspace{-12pt}%
  \centering
  \includegraphics[width=\linewidth]{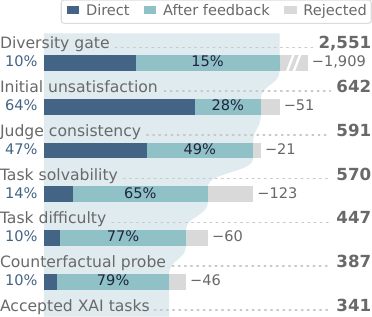}
  \caption{Progression of XAI task candidates. Direct denotes sketches that pass on the first attempt, while After feedback denotes sketches that pass after at least one feedback-driven revision at any stage of the whole synthesis, thus the funnel is monotonic.}
  \label{fig:quality_gates}
  \vspace{-12pt}%
\end{wrapfigure}

\myparagraph[par:method:rollouts-judgments]{Rollouts \& Judgments.} After generating a task and its verifier, %
we use heterogeneous agents to attempt each task: successful rollouts provide evidence of solvability, success rates characterize empirical difficulty, and their trajectories and outcomes help validate the verifier.
Specifically, we collect six rollouts per task, half with hints, from 18 model--harness pairs formed by three open-source models---GLM-5.2~\citep{glm5}, DeepSeek-V4-Flash~\citep{deepseekv4}, and MiniMax-M3~\citep{minimaxm3}---and six harnesses, comprising three coding agents (OpenCode~\citep{opencode2025}, Codex~\citep{codex}, Claude Code~\citep{claudecode}) and three claw agents (Nanobot~\citep{nanobot}, OpenClaw~\citep{openclaw}, Hermes Agent~\citep{hermesagent}).

We further validate each rollout through three levels of judgment, which we view as test-time scaling for verification: %
a verifier $V$ the generator just synthesized; %
a LLM judge that semantically assesses the task, rubric, trajectory, and verifier results; 
an agent judge using OpenCode that actively inspects the initial and final environments in multi turns~\citep{zhuge2024agent}. 
Although agent judge can be costly, we only use it during synthesis to cross-check judgments and identify cases requiring verifier revision. The synthesized programmatic verifier is finally used as the reward function during RL.

\myparagraph[par:method:quality-gates]{Quality Gates.}
The collected rollouts and judgments provide execution evidence but do not by themselves establish task validity. We propose five quality gates to assess task solvability, difficulty, and verifier reliability, accepting candidates that pass all gates.

\begin{itemize}[leftmargin=0.5cm, nosep]
    \item \textbf{Judge Consistency:} For rollout $j$, let $W_j$ and $a_j$ denote its final state and answer, and let $V_{\text{LLM}}^{(j)}$ and $V_{\text{agent}}^{(j)}$ denote the LLM and agent judges' verdicts. The judge consistency gate requires $V(W_0,W_j,a_j)=V_{\text{LLM}}^{(j)}=V_{\text{agent}}^{(j)}$ for every $j\in\{1,\ldots,6\}$. Disagreements and the associated evidence will be returned to the generator to revise the rubric $R$ and verifier scripts $V$.
    \item \textbf{Initial Unsatisfaction:} This gate requires $V(W_0,W_0,\emptyset)=0$, where $\emptyset$ denotes the absence of any agent action, rejecting tasks already considered complete in the initial state.
    \item \textbf{Task Solvability:} Agreement among judges does not establish solvability if all rollouts fail. The task solvability gate requires at least one successful rollout, $\sum_{j=1}^{6}V(W_0,W_j,a_j)\geq1$. Each successful trajectory must also use every interface declared during task generation, providing execution evidence of the intended cross-interface workflow.
    \item \textbf{Task Difficulty:} The task difficulty gate estimates empirical difficulty using $k=\sum_{j\in\mathcal{U}}V(W_0,W_j,a_j)$, where $\mathcal{U}$ indexes the three rollouts without hint. Tasks with $k\leq1$ are labeled \textit{hard}, those with $k=2$ are labeled \textit{medium}, and those with $k=3$ are labeled \textit{too easy} which will be returned to the generator to increase  difficulty.
    \item \textbf{Counterfactual Probe:} The preceding gates rely partly on model-based judgments, including agent judge's feedback for revising the verifier $V$. Given LLMs' susceptibility to sycophancy~\citep{sharma2024towards}, we add a counterfactual probe to test the stability of this feedback under incorrect outcome claims. Specifically, given a rollout with verdict $V(W_0,W,a)$, we present the agent judge with the counterfactual claim $1-V(W_0,W,a)$ and ask it to reinvestigate the execution evidence. Let $V_{\text{probe}}\in\{0,1\}$ denote the judge's verdict after reinvestigation. The candidate passes only if $V_{\text{probe}}=V(W_0,W,a)$; otherwise, it fails.
\end{itemize}

\definecolor{ifcli}{HTML}{2F5C8F}   %
\definecolor{ifmcp}{HTML}{1A7F37}   %
\definecolor{ifgui}{HTML}{C2620A}   %
\definecolor{ifclibg}{HTML}{E8EFF8}
\definecolor{ifmcpbg}{HTML}{E6F3EA}
\definecolor{ifguibg}{HTML}{FBEEE1}
\definecolor{casepromptbg}{HTML}{F4F5F7}

\providecommand{\ifacebadge}[2]{%
  \colorbox{#1}{\makebox[2.1em][c]{\textcolor{white}{\sffamily\bfseries\scriptsize #2}}}}
\providecommand{\bCLI}{\ifacebadge{ifcli}{CLI}}
\providecommand{\bMCP}{\ifacebadge{ifmcp}{MCP}}
\providecommand{\bGUI}{\ifacebadge{ifgui}{GUI}}
\providecommand{\hoparrow}{\,\textcolor{gray}{$\rightarrow$}\,}

\begin{table}[t]
  \caption{Case study of XAI task. Only the shaded block is shown to the solver, and the remaining blocks are internal to the pipeline. The interface flow MCP \hoparrow CLI\hoparrow GUI\hoparrow CLI\hoparrow MCP{} consists of five sub-tasks and returns to both CLI and MCP after the GUI write, so the task cannot be completed within a single interface. Each sub-task that changes the world state is checked by one deterministic verifier script, and task success requires all verifier scripts to pass.}
  \label{tab:task_case_study}
  \centering
  \scriptsize
  \setlength{\tabcolsep}{3pt}
  \renewcommand{\arraystretch}{1.15}
  \begin{tabularx}{\textwidth}{@{}c@{\hspace{4pt}}l>{\raggedright\arraybackslash}X l@{}}
    \toprule
    \multicolumn{4}{@{}l@{}}{\textbf{User request} \;\small\textnormal{(\texttt{task.md})}} \\
    \midrule
    \rowcolor{casepromptbg}
    \multicolumn{4}{@{}p{\textwidth}@{}}{%
      \itshape
      Marco gave the green light in our chat --- ``Go, order it today.'' The Penalty Box rig goes live
      Apr~11 and the rack PDUs still need the production surge units locked before Brendan starts the swap.
      \par\smallskip
      I need to write and run a script that parses the rack inventory
      (\texttt{Documents/northstarav/inventory.csv}), the Penalty Box bill of materials
      (\texttt{Documents/northstarav/jobs/penalty-box-display-wall.md}), and the APC spec sheet
      (\texttt{Downloads/apc-surge-specsheet.pdf}), cross-references stock against the BOM, extracts each
      surge candidate's specs (outlets, joules, cord length, trip-indicator presence), ranks them by spec
      fit to the rack requirement, and saves the ranked shortlist to
      \texttt{./penalty-box-rack-unit-audit.csv} with columns for product id, outlet count, joules, cord
      length, trip-indicator flag, and quantity needed.
      \par\smallskip
      The top-ranked unit needs to go in the Amazon cart --- sign in with the credentials in
      \texttt{Documents/personal/credentials.md}, find the surge protector matching the top spec (two
      candidates share the same outlets, joules, and cord length in the listing details --- pick the one
      whose visible design has a white body and a digital display panel), add it to the cart, create a
      restock note at \texttt{Documents/Notes/penalty-box-surge-restock.md} with the locked unit and spec
      match, and send a direct message to Brendan letting him know which surge unit is locked and that
      it's queued for the PDU swap.\strut} \\
    \midrule
    \multicolumn{4}{@{}l@{}}{\textbf{Interface flow} \;\small\textnormal{(\texttt{mcp>cli>web>cli>mcp}, five sub-tasks that revisit CLI and MCP)}} \\
    \midrule
    \# & Interface & Sub-task and grounded world state & Verifier \\
    \cmidrule(r){1-1}\cmidrule(lr){2-2}\cmidrule(lr){3-3}\cmidrule(l){4-4}
    \rowcolor{ifmcpbg}
    1 & \bMCP & Read the client thread and confirm the go-ahead to order.
        \newline \textcolor{ifmcp}{$\vartriangleright$~reads} \texttt{direct\_messages} thread~1, msg~13 (Marco Delgado:
        ``Go~--~4x QN90C, bezel's fine. Order it today.'') & \textit{read-only} \\
    \rowcolor{ifclibg}
    2 & \bCLI & Write and run a script that joins three heterogeneous local sources, ranks the surge
        candidates by spec fit, and emits the shortlist.
        \newline \textcolor{ifcli}{$\vartriangleright$~reads} \texttt{inventory.csv} (CSV),
        \texttt{penalty-box-display-wall.md} (BOM), \texttt{apc-surge-specsheet.pdf} (PDF)
        \newline \textcolor{ifcli}{$\vartriangleright$~writes} \texttt{penalty-box-rack-unit-audit.csv}
        & \texttt{check\_audit.py} \\
    \rowcolor{ifguibg}
    3 & \bGUI & Sign in to the storefront and cart the top-ranked unit. Image-decisive: APC~p151
        (white body, LCD timer panel) vs.\ GE~p155 (black body, no LCD) are identical on every listed
        spec, so the choice is only resolvable from the product photo.
        \newline \textcolor{ifgui}{$\vartriangleright$~writes} \texttt{amazon\_store.cart\_items}, product resolved by
        slug \texttt{apc-performance-surgearrest-7-outlet} & \texttt{check\_cart.py} \\
    \rowcolor{ifclibg}
    4 & \bCLI & Fold the carted unit and the spec match into a restock note.
        \newline \textcolor{ifcli}{$\vartriangleright$~writes} \texttt{Documents/Notes/penalty-box-surge-restock.md}
        & \texttt{check\_note.py} \\
    \rowcolor{ifmcpbg}
    5 & \bMCP & Notify the install crew which unit is locked and that it is queued for the PDU swap.
        \newline \textcolor{ifmcp}{$\vartriangleright$~writes} \texttt{direct\_messages} thread~4 (Brendan Olsrud, resolved
        by participant name) & \texttt{check\_dm.py} \\
    \midrule
    \multicolumn{4}{@{}l@{}}{\textbf{Quality-gate evidence}} \\
    \midrule
    \multicolumn{4}{@{}p{\textwidth}@{}}{%
      \begin{minipage}[t]{0.485\textwidth}\scriptsize\raggedright
        \textbf{Judge consistency.} $V\!=\!V_{\text{LLM}}\!=\!V_{\text{agent}}$ on 6/6 rollouts. \\
        \textbf{Initial unsatisfaction.} $V(W_0,W_0,\emptyset)\!=\!0$, as all four scripts fail on the
        untouched world. \\
        \textbf{Task solvability.} 1/3 hinted rollouts completed.\strut
      \end{minipage}\hfill
      \begin{minipage}[t]{0.485\textwidth}\scriptsize\raggedright
        \textbf{Task difficulty.} $k\!=\!0$ of 3 unhinted rollouts succeeded $\Rightarrow$ \textit{hard}. \\
        \textbf{Counterfactual probe.} The agent judge defended its verdict under
        the opposite claim and edited no verifier file.\strut
      \end{minipage}} \\
    \bottomrule
  \end{tabularx}
\end{table}

\begin{table}[t]
  \centering
  \caption{XAI task statistics. SR: mean success rate over synthesis rollouts from a pool of 18 agent configs; fixes: feedback-driven revisions per task at each gate. Task execution and interface usage are measured by GLM-5.2 with OpenCode. File operations include reading, editing, and creation.}
  \label{tab:task_syn_basics}
  \scriptsize
  \begin{adjustbox}{max width=\linewidth}
    \begin{tabular}{@{}lll@{}}
      \begin{tabular}[t]{@{}lr@{}}
        \toprule
        \multicolumn{2}{c}{\itshape Task corpus} \\
        \midrule
        Tasks & 341 \\
        Hard tasks & 273 \\
        Medium tasks & 68 \\
        Tasks per persona & 9.5 \\
        Tasks per world & 0.53 \\
        Interface switches / task & 3.1 \\
        \midrule
        \multicolumn{2}{c}{\itshape Success rate during Synthesis} \\
        \midrule
        SR w/o hint & 21.3\% \\
        SR w/ hint & 45.6\% \\
        \bottomrule
      \end{tabular}
      &
      \begin{tabular}[t]{@{}lrrr@{}}
        \toprule
        \itshape Synthesis Revisions & Mean & Median & P90 \\
        \midrule
        Initial-unsat.\ fixes & 0.05 & 0 & 0 \\
        Judge-consist.\ fixes & 1.09 & 1 & 3 \\
        Solvability fixes & 1.16 & 1 & 3 \\
        Difficulty fixes & 0.39 & 0 & 1 \\
        Probe fixes & 0.18 & 0 & 1 \\
        \midrule
        \itshape Costs per task & Mean & Median & P90 \\
        \midrule
        Revision rounds & 2.5 & 2 & 5 \\
        Generator tokens in (M) & 69.8 & 58.3 & 130.9 \\
        Generator tokens out (M) & 1.1 & 1.0 & 2.0 \\
        \bottomrule
      \end{tabular}
      &
      \begin{tabular}[t]{@{}lrrr@{}}
        \toprule
        \itshape Task Execution & Mean & Median & P90 \\
        \midrule
        LLM calls & 80 & 74 & 121 \\
        Tool calls & 81 & 76 & 124 \\
        LLM in tokens (M) & 4.1 & 3.7 & 6.5 \\
        LLM out tokens (K) & 20.6 & 18.0 & 34.7 \\
        \midrule
        \itshape Interface Usage & Mean & Median & P90 \\
        \midrule
        GUI websites visited & 1.1 & 1.0 & 2.0 \\
        MCP servers connected & 2.4 & 2.0 & 3.0 \\
        Local files processed & 5.0 & 4.0 & 9.0 \\
        \bottomrule
      \end{tabular}
    \end{tabular}
  \end{adjustbox}
\end{table}

\myparagraph[par:method:case-study]{Case Study.}
Table~\ref{tab:task_case_study} shows a case study of an accepted task that illustrates these properties. The task asks an audio-visual technician commissioning a display wall for a sports bar to lock the production surge units for the rack before the install crew starts the swap. The owner's approval sits in a chat thread from an earlier day of the world, and the workspace holds a rack inventory, a bill of materials, and a product specification sheet accumulated since day 0, so solving the task depends on state built up over the world's history. The agent first confirms the go-ahead in the direct-message MCP service, then writes and runs a script via CLI in the terminal  that joins the three local sources, ranks the candidate units by their specifications, and saves the shortlist. The GUI step happens on the storefront, where two candidates match on every listed specification and differ only in their visible design, so the agent must inspect the product images to identify the correct one. The bottom block of Table~\ref{tab:task_case_study} reports the evidence collected at the five quality gates. The three levels of judgment agree on all six rollouts, all four checks fail on the initial state, one of the three hinted rollouts completes the task, and none of the three unhinted rollouts does, so the task is classified as hard.  Under the counterfactual probe, the agent judge insists on its original verdict. See more examples in Appendix~\ref{appendix:case_study}.

In summary, candidates that fail any gate are returned to the generator with diagnostic evidence for up to two revision rounds, after which they must re-pass all gates.
Accepted tasks are added to the diversity pool, while failed ones provide lessons to be absorbed into  generator's evolving skills.
Table~\ref{tab:task_syn_basics} and Figure~\ref{fig:quality_gates} summarize synthesis statistics and candidate progression, with stage-by-stage details and prompts in Appendix~\ref{appendix:task_synth}. During rollouts, XAI tasks have a mean success rate of only 21.3\%, and even GLM-5.2 sometimes requires more than 100 steps to complete a task. Consistent with the solvability and difficulty gates, every accepted task is solved by at least one agent, while none is solved by all agents.
XAI tasks consists of $4.2$ sub-tasks and $3.1$ interface switches on average, and 74.5\% revisit a previously used interface, requiring agents to preserve state across interface transitions rather than completing the task within a single interface.

\subsection{General CUA Training}
\label{sec:method:training}
We train general CUAs in our synthesized worlds using GRPO~\citep{grpo}, with a validated verifier that provides dense but outcome-based rewards. For rollout $j$ of task $\tau$, let $(W_j,a_j)$ denote its final world state and answer, and let $e_j\in\{0,1\}$ indicate a policy format failure (e.g., invalid tool call). We assign $r_j=-1$ when $e_j=1$. For $e_j=0$, the reward combines with checklist scores $v_i$:
\begin{equation}
    \small
  r_j = \min\left\{1,\,
  V(W_0,W_j,a_j)
  + \frac{0.1}{N}\sum_{i=1}^{N}v_i(W_0,W_j,a_j)
  \right\},
  \label{eq:training_reward}
\end{equation}
where $N$ is the number of checklist criteria.
Thus, successful tasks receive unit reward, while checklist scores provide partial credit for incomplete trajectories, yielding denser signals for long-horizon learning.
However, small models often produce zero-variance reward groups, providing no GRPO learning signal. Inspired by DAPO~\citep{dapo}, we maintain for each task $\tau$ an exponential moving estimate~\citep{roberts2000control} of the required rollout multiplier $\lceil K/G\rceil$, where $K$ is the rollout pool size and $G$ the target group size. The multiplier is initialized at 1, capped at 3, and updated with weight 0.8 after each optimization step.
When $K>G$, we select a group from  possible $\binom{K}{G}$ groups containing samples from both the lower and upper reward halves to encourage reward diversity.
Let $\mu_\tau$ and $\sigma_\tau$ denote the mean and population standard deviation of rewards in a completed rollout pool of size $K$. Then, the trajectory advantage is defined as
\begin{equation}
\small
A_j =
\begin{cases}
(r_j-\mu_\tau)/\sigma_\tau, & \sigma_\tau>0,\\
-K^{-1/2}, & \sigma_\tau=0 \text{ and } e_j=1\ \forall j,\\
0, & \text{otherwise},
\end{cases}
\label{eq:advantage}
\end{equation}
where the second case is designed to provide format-related learning signal even when all rollouts receive the same format error reward, which is common for long-horizon rollouts during the early stages of training.
Note that the rollout will be terminated once a format error occurs, and gradients are restricted to the generated tokens in the failing step. More details are provided in Appendix~\ref{appendix:training}.

\section{Experiments}

\myparagraph[par:experiments:dataset-benchmarks]{Dataset \& Benchmark.} 
We synthesized 341 XAI tasks by \ours, and we randomly split them into 100 for SFT, 208 for RL, and 33 for in-domain evaluation. 
To evaluate the OOD generalization of our agent, we further include 8 benchmarks across 3 categories of agent tasks:
WebArena (WA)~\citep{zhou2024webarena} and Online-Mind2Web (O2W)~\citep{onlinemind2web} for web agents; QwenClawBench (QCB)~\citep{qwenclawbench} and ClawEval (CLE)~\citep{claweval} for claw agents; CUA-Gym (CUG)~\citep{wang2026cuagymscalingverifiabletraining}, OSWorld-Verified (OSV)~\citep{xie2024osworld}, OSWorld 2.0 (OS-2)~\citep{yuan2026osworld2}, and Agents' Last Exam (ALE)~\citep{agentslastexam} for computer-use agents.

\myparagraph[par:experiments:models-baselines]{Model \& Baselines.}
We use two base models at different scales for training: Qwen3.6-35B-A3B~\citep{qwen36_35b_a3b} and Qwen3.5-4B~\citep{qwen35blog}. 
We compare our trained models against a range of frontier models, including GPT-6-Astra~\citep{gpt6}, GPT-5.6-Sol~\citep{gpt56}, GPT-5.5~\citep{gpt55}, Claude Opus 5~\citep{opus5}, Claude Opus 4.8~\citep{opus48}, and GLM-5.2~\citep{glm5}.
We also include numbers reported in prior computer-use agent work for reference, including CUA-Gym~\citep{wang2026cuagymscalingverifiabletraining} and OpenForgeRL~\citep{yu2026openforgerltrainharnessnativeagents}.\footnote{For reference only, due to their unavailable checkpoints or evaluation with traditional GUI harnesses.}

\myparagraph[par:experiments:implementation-details]{Implementation Details.}
We use 8 NVIDIA B200 GPUs for both synthesis and training.
During synthesis, we use GLM-5.2 as both the generator and the agent-as-a-judge with OpenCode harness~\citep{opencode2025}. %
For evaluation, we use OpenCode equipped with GUI (e.g., \texttt{click} and \texttt{type}) and MCP (e.g., Playwright) tools, and enable high reasoning effort for the model.
The maximum number of agent steps is set to 256 for most benchmarks and 512 for OSWorld 2.0 and ALE.
The context window is set to 100K tokens with automatic context compaction. 
For training, we first construct a seed dataset by collecting 100 GLM-5.2 trajectories on the SFT split, and warm-start the Qwen base models with SFT for 3 epochs.
We then perform RL for up to 64 training steps, using a batch size of 8 tasks and $G=4$ rollouts per task.
Environment management and RL training are built on Orchard~\citep{peng2026orchard} and slime~\citep{slime_github}. More details are in Appendix~\ref{appendix:impl}.

\begin{table}[t]
  \definecolor{deltagain}{HTML}{1A7F37}
  \newcommand{\avgdelta}[1]{\rlap{\textsubscript{\textcolor{deltagain}{#1}}}}
  \newcommand{\cw}[1]{\makebox[20.25pt][l]{#1}}
  \newcommand{\hsp}{\hspace{0pt plus 1fil}}
  \newcommand{\hdr}[1]{\hbox to 20pt{#1}}
  \caption{Main results on OOD benchmarks. We report pass rates for O2W, WA, and CLE, and mean score for the other benchmarks.
  $^{\dagger}$ Numbers reported in the official papers.}
  \label{tab:main_results}
  \centering
  \scriptsize
  \setlength{\tabcolsep}{4.75pt}
  \begin{adjustbox}{max width=\linewidth}
  \begin{tabular}{@{}l@{\hspace{7pt}}l@{\hspace{7pt}}cccccccc@{\hspace{6pt}}|c@{\hspace{14pt}}}
    \toprule
     & & \multicolumn{2}{c}{Web Agent} & \multicolumn{2}{c}{Claw Agent} & \multicolumn{4}{c@{\hspace{6pt}}}{Computer-Use Agent} & \\
    \cmidrule(lr){3-4} \cmidrule(lr){5-6} \cmidrule(lr){7-10}
    Model & Harness & \cw{\hdr{O\hsp 2\hsp W}} & \cw{\hbox to 20pt{\hfil WA\hfil}} & \cw{\hdr{Q\hsp C\hsp B}} & \cw{\hdr{C\hsp L\hsp E}} & \cw{\hdr{C\hsp U\hsp G}} & \cw{\hdr{O\hsp S\hsp V}} & \cw{\hdr{OS\hsp-\hsp 2}} & \cw{\hdr{A\hsp L\hsp E}} & Avg. \\
    \midrule
    \multicolumn{11}{l}{\itshape Frontier Models} \\
    \midrule
    GLM-5.2 & OpenCode & \cw{86.67} & \cw{67.26} & \cw{64.44} & \cw{59.67} & \cw{78.78} & \cw{62.46} & \cw{45.00} & \cw{41.39} & 63.21 \\
    Opus-4.8 & OpenCode & \cw{91.01} & \cw{71.24} & \cw{64.74} & \cw{76.33} & \cw{\textbf{84.52}} & \cw{68.58} & \cw{48.32} & \cw{46.11} & 68.86 \\
    Opus-5 & OpenCode & \cw{\textbf{96.60}} & \cw{\textbf{75.66}} & \cw{55.02} & \cw{77.67} & \cw{84.39} & \cw{\textbf{74.64}} & \cw{52.79} & \cw{49.85} & 70.83 \\
    GPT-5.5 & OpenCode & \cw{72.22} & \cw{70.80} & \cw{62.87} & \cw{63.67} & \cw{82.71} & \cw{61.46} & \cw{47.12} & \cw{52.39} & 64.16 \\
    GPT-5.6-Sol & OpenCode & \cw{86.72} & \cw{67.70} & \cw{61.34} & \cw{64.33} & \cw{82.83} & \cw{58.48} & \cw{46.89} & \cw{53.45} & 65.22 \\
    GPT-6-Astra & OpenCode & \cw{92.78} & \cw{74.34} & \cw{\textbf{65.21}} & \cw{\textbf{80.33}} & \cw{83.56} & \cw{71.28} & \cw{\textbf{60.77}} & \cw{\textbf{56.58}} & \textbf{73.11} \\
    \midrule
    \multicolumn{11}{l}{\itshape 4B-Level Models}\\
    \midrule
    Qwen3.5 & OpenCode & \cw{26.39} & \cw{30.09} & \cw{44.06} & \cw{26.67} & \cw{21.39} & \cw{18.18} & \cw{0.87} & \cw{7.00} & 21.83 \\
    \textbf{\ours} & OpenCode & \cw{\textbf{45.65}} & \cw{\textbf{39.82}} & \cw{\textbf{48.37}} & \cw{\textbf{36.67}} & \cw{\textbf{37.35}} & \cw{\textbf{22.48}} & \cw{\textbf{1.03}} & \cw{\textbf{8.96}} & \textbf{30.04}\avgdelta{+8.21} \\
    \midrule
    \multicolumn{11}{l}{\itshape 35B-A3B-Level Models} \\
    \midrule
    Qwen3.6 & OpenCode & \cw{69.60} & \cw{53.54} & \cw{57.86} & \cw{38.00} & \cw{53.73} & \cw{33.16} & \cw{13.22} & \cw{23.57} & 42.84 \\
    OpenForgeRL$^{\dagger}$ & OpenClaw & \cw{--} & \cw{--} & \cw{33.7} & \cw{45.1} & \cw{--} & \cw{--} & \cw{--} & \cw{--} & -- \\
    CUA-Gym$^{\dagger}$ & R-GUI & \cw{--} & \cw{44.5} & \cw{--} & \cw{--} & \cw{--} & \cw{62.1} & \cw{--} & \cw{--} & -- \\
    \textbf{\ours} & OpenCode & \cw{\textbf{80.77}} & \cw{\textbf{67.26}} & \cw{\textbf{61.44}} & \cw{\textbf{54.67}} & \cw{\textbf{65.95}} & \cw{\textbf{49.22}} & \cw{\textbf{21.73}} & \cw{\textbf{26.31}} & \textbf{53.42}\avgdelta{+10.58} \\
    \bottomrule
  \end{tabular}
  \end{adjustbox}
\end{table}

\myparagraph{Main Results.}
Table~\ref{tab:main_results} summarizes the main results. We highlight three key findings.
(1) \textit{Broad OOD generalization.} Training exclusively in \ours's synthesized worlds improves both model scales on all eight benchmarks, spanning web, claw, and computer-use agents. The 35B-A3B and 4B models gain $+10.58$ and $+8.21$ points on average, respectively, demonstrating broad transfer across unseen tasks and scenarios.
(2) \textit{Larger models benefit more.} The 35B-A3B model improves more than the 4B model on average ($+10.58$ vs.\ $+8.21$), with particularly large gains on CLE ($+16.67$), OSV ($+16.06$), WebArena ($+13.72$), and CUG ($+12.22$). This suggests that the benefits of \ours's long-horizon, cross-interface supervision persist, and may even increase, at larger model capacity.
(3) \textit{Narrowing the gap to frontier models.}
\ours raises the 35B-A3B model's average score from $42.84$ to $53.42$, substantially reducing its gap to the listed frontier models.
The resulting model further matches GLM-5.2 on WebArena ($67.26$) and exceeds GPT-5.5 on O2W ($80.77$ vs.\ $72.22$).

\section{Analysis}

\begin{figure}[t]
  \centering
  \includegraphics[max width=0.9\linewidth]{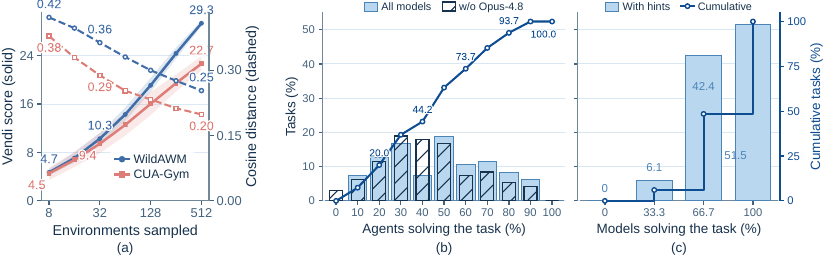}%
  \begin{subcaptiongroup}%
    \phantomsubcaption\label{fig:env_diversity}%
    \phantomsubcaption\label{fig:task_solvability_with_opus}%
    \phantomsubcaption\label{fig:task_solvability_hints}%
  \end{subcaptiongroup}
  \caption{(a) Environment diversity using Qwen3-Embedding-8B~\citep{qwen3embedding} representations of environment artifacts. Shaded bands are 95\% intervals of the Vendi score~\citep{friedman2023vendi}. (b) Task solvability across 24 model--harness configurations: the three models and six harnesses in \parref{par:method:rollouts-judgments}, plus Opus-4.8. (c) Solvability with hints w/o Opus-4.8.}
  \label{fig:task_solvability}
\end{figure}

\subsection{Synthesis Quality}
\myparagraph[][]{Environment.}
A \ours's environment state accumulates a persona's activities over multiple days, with a median size of 52.4K tokens, while CUA-Gym uses task-specific injected states with a median size of only 1.1K tokens.
Despite its substantially larger environment scale, \ours is also more diverse.
As shown in Figure~\ref{fig:env_diversity}, we find that \ours achieves consistently higher Vendi score across all sample sizes. Its higher nearest-neighbor cosine distance (0.25 vs.\ 0.20) further indicates lower redundancy.
Notably, we achieve these gains using only open-source models, whereas most existing work relies on proprietary frontier models for synthesis.

\begin{figure}[t]
  \centering
  \includegraphics[width=0.9\textwidth]{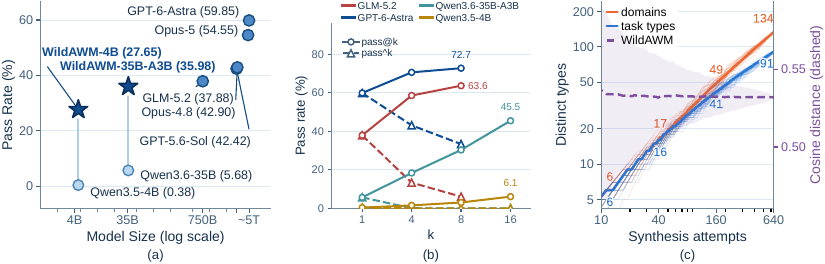}%
  \begin{subcaptiongroup}%
    \phantomsubcaption\label{fig:wildawm_performance}%
    \phantomsubcaption\label{fig:valid_pass_k}%
    \phantomsubcaption\label{fig:task_scaling}%
  \end{subcaptiongroup}
  \vspace{-5pt}
  \caption{(a)/(b) Performance on \ours validation set, and proprietary model sizes are estimated. (c) Rarefaction of the 341 accepted tasks by LLM-classified task types and domains.} 
  \label{fig:wildawm_scaling_trio}
\end{figure}

\myparagraph[][]{Task \& Verifier.}
We answer these important task synthesis questions:
(1) \textbf{Are XAI tasks solvable?}
Rather than relying on a single strong model as evidence of solvability, we evaluate each task with 24 heterogeneous agents (model--harness pairs). As shown in Figure~\ref{fig:task_solvability_with_opus}, every task is solved by at least one agent, while the three open-source models alone used during synthesis already solve over 95\% of tasks. Figure~\ref{fig:task_solvability_hints} further evaluates our hints using only these open-source models. With hints, every task is solved by at least one of the three models, and 51.5\% by all three, providing stronger evidence that XAI tasks are solvable through executable paths.
(2) \textbf{Are XAI tasks challenging, even for frontier models?} We evaluate frontier and open-source models on synthesized tasks to assess task difficulty in Figure~\ref{fig:wildawm_performance}, and find that GPT-6-Astra achieves only about 60\% \passatk{1}, while our trained agents approach the performance of GLM-5.2 with substantially fewer parameters.
Figure~\ref{fig:valid_pass_k} shows that difficulty persists under test-time scaling: even after eight attempts, GPT-6-Astra still fails on 27.3\% of tasks and \passtok drops rapidly, indicating that open-source-synthesized tasks remain challenging enough and could even provide useful training signals for frontier models.
(3) \textbf{Are synthesized verifiers reliable?} We compare our verifier against multiple agent-as-a-judge models on 100 randomly sampled rollouts in Table~\ref{tab:verifier_reliability}.
Our verifier achieves 95\% agreement with the majority, with a Cohen's $\kappa$ of 0.90, suggesting that our verifier is reliable within natural variation of stronger but heavier agent-as-a-judge judgments.
(4) \textbf{Is the synthesis pipeline scalable?}
As shown in Figure~\ref{fig:task_scaling}, synthesis continues to expand task coverage, reaching 134 domains and 91 task types at a 53\% acceptance rate without clear saturation.
Meanwhile, mean pairwise cosine distance remains stable as the pool grows, suggesting that increased coverage does not come from increasingly similar tasks.

\begin{table*}[t]
    \centering

    \begin{minipage}[t]{0.5\textwidth}
        \centering
        \caption{Study on verifier reliability.
        Cons.: the fraction of rollouts receiving the same verdict from both sides. $\kappa$: Cohen's kappa.}
        \label{tab:verifier_reliability}

        \resizebox{\linewidth}{!}{
        \begin{tabular}{lccccc}
            \toprule
            Agent Judge\protect\footnotemark & Cons. & Precision & Recall & F1 & $\kappa$ \\
            \midrule
            GLM-5.2         & 0.95 & 0.98 & 0.93 & 0.95 & 0.90 \\
            GPT-5.6-Sol     & 0.92 & 0.85 & 0.98 & 0.91 & 0.83 \\
            Claude-Opus-5   & 0.92 & 0.98 & 0.88 & 0.93 & 0.84 \\
            Judge majority  & 0.95 & 0.98 & 0.93 & 0.95 & 0.90 \\
            \midrule
            \multicolumn{6}{l}{Inter-Judge Agreement} \\
            \midrule
            GLM-5.2 vs.\ Opus-5       & 0.97 & 1.00 & 0.95 & 0.97 & 0.94 \\
            GPT-5.6-Sol vs.\ Opus-5   & 0.89 & 1.00 & 0.80 & 0.89 & 0.79 \\
            \bottomrule
        \end{tabular}
        }
    \end{minipage}
    \hfill
    \begin{minipage}[t]{0.47\textwidth}
        \centering
        \caption{Ablation of quality gates on four benchmarks, reported as success rate (\%).
        Val.: our synthesized in-domain validation set.}
        \label{tab:ablation_study}

        \resizebox{\linewidth}{!}{
        \begin{tabular}{lccccc}
            \toprule
            Model & O2W & QCB & OSV & Val. & Avg. \\
            \midrule
            Qwen-3.5-4B
                & 26.39 & 44.06 & 18.18 & 0.38 & 22.25 \\
            \textbf{\ours + SFT}
                & 38.49 & \textbf{50.17} & 17.17 & \textbf{21.59} & \textbf{31.86} \\
            \midrule
            w/o Init.\ Unsat.
                & 33.21 & 42.36 & 17.71 & 3.05 & 24.08 \\
            w/o Judge Cons.
                & 29.10 & 37.72 & 19.62 & 3.79 & 22.56 \\
            w/o Task Solv.
                & \textbf{44.87} & 44.15 & 12.12 & 15.91 & 29.26 \\
            w/o Task Diff.
                & 39.05 & 42.29 & 16.54 & 7.58 & 26.37 \\
            w/o CF Probe
                & 31.37 & 42.55 & \textbf{24.09} & 13.64 & 27.91 \\
            \bottomrule
        \end{tabular}
        }
    \end{minipage}

\end{table*}
\footnotetext{Each judge model represents multiple runs with different harnesses. GLM-5.2: OpenCode, Claude Code, and Codex; GPT-5.6-Sol: OpenCode and Codex; Claude-Opus-5: OpenCode and Claude Code.}

\begin{table}[t]
  \centering
  \caption{Ablations on RL training with Qwen3.6-35B-A3B as the base model.
  }
  \label{tab:ablation_on_rl}
  \scriptsize
  \definecolor{deltadrop}{HTML}{C0392B}
  \newcommand{\rlavgdrop}[1]{\textsubscript{\textcolor{deltadrop}{\ensuremath{#1}}}}
  \begin{adjustbox}{max width=\linewidth}
    \begin{tabular}{@{}lrrrrrrrrl@{}}
      \toprule
      & \multicolumn{2}{c}{Web Agent} & \multicolumn{2}{c}{Claw Agent} & \multicolumn{4}{c}{Computer-Use Agent} & \\
      \cmidrule(lr){2-3}\cmidrule(lr){4-5}\cmidrule(lr){6-9}
      Method & O2W & WA & QCB & CLE & CUG & OSV & OS-2 & ALE & Avg. \\
      \midrule
      \textbf{\ours} & \textbf{80.77} & \textbf{67.26} & 61.44 & 54.67 & 65.95 & \textbf{49.22} & \textbf{21.73} & 26.31 & \textbf{53.42} \\
      w/o Format Reward & 77.91 & 58.85 & \textbf{62.17} & \textbf{61.00} & 66.43 & 40.83 & 19.31 & \textbf{30.93} & 52.18\rlavgdrop{-1.24} \\
      w/o Checklist Reward & 73.38 & 64.89 & 55.06 & 59.18 & 66.53 & 44.29 & 18.51 & 27.51 & 51.17\rlavgdrop{-2.25} \\
      w/o SFT Warm Start & 66.29 & 57.96 & 59.74 & 51.67 & \textbf{68.34} & 42.78 & 13.54 & 24.59 & 48.11\rlavgdrop{-5.31} \\
      \bottomrule
    \end{tabular}
  \end{adjustbox}
\end{table}

\subsection{Ablations on Quality Gates}
To isolate the contribution of each synthesis gate, we synthesize SFT data with one gate removed at a time and train Qwen3.5-4B model under an identical settings. 
As shown in Table~\ref{tab:ablation_study}, the full quality-control strategy achieves the best overall performance,  while removing any gate reduces the average by 2.60--9.30 points. 
Judge Consistency and Initial Unsatisfaction are particularly important: removing them drops the average to 22.56 and 24.08, respectively, and reduces performance on in-domain validation set from 21.59 to only 3.79 and 3.05. 
Task Difficulty also contributes substantially, with its removal decreasing the validation score to 7.58.  
These results show that five gates are complementary: together they produce more reliable training data and substantially improve both in-domain task learning and generalization performance on OOD benchmarks.

\subsection{Ablations on RL}
We ablate key components of our RL recipe to measure their impact in Table~\ref{tab:ablation_on_rl}, and find that removing format penalty cause significant drop in WA and OSV, while its higher CLE pass rate is not matched by a higher CLE mean score (69.89 vs.\ 69.45 \ours), i.e., more tasks end just above the pass threshold.
We conjecture that the format reward is particularly important for GUI agents, where actions must be correctly parsed to interact with visual interfaces. 
Without checklist partial credit, performance decreases on most benchmarks, indicating that the denser reward improves credit assignment. Notably, removing the warm start causes the largest degradation, highlighting the importance of a strong initialization for effective RL training at this small model scale.

\begin{figure}[t]
  \centering
  \includegraphics[max width=0.88\linewidth]{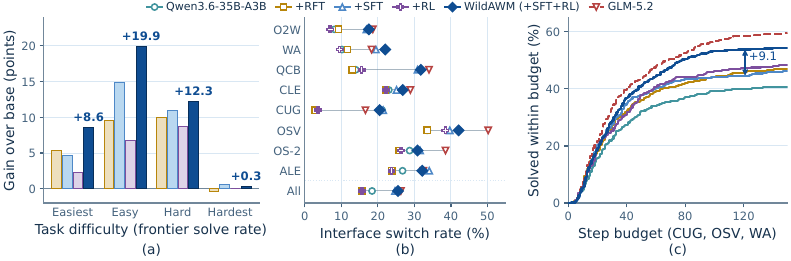}%
  \begin{subcaptiongroup}%
    \phantomsubcaption\label{fig:behavior_difficulty}%
    \phantomsubcaption\label{fig:behavior_interleaving}%
    \phantomsubcaption\label{fig:behavior_step_budget}%
  \end{subcaptiongroup}
  \vspace{-5pt}
  \caption{Behavior analysis of base model, \ours and its variants (+RFT, +SFT, and +RL, i.e., RL without SFT), and GLM-5.2. (a) Pass-rate gain over the base model on OOD tasks, split into four equal quartiles by how many of frontier models solve each task. (b) Fraction of consecutive tool calls that switch between interface families (e.g., GUI $\rightarrow$ CLI). (c) Tasks solved within a step budget.}
  \label{fig:model_behavior}
\end{figure}

\subsection{Policy Behavior Analysis}
\label{sec:analysis:behavior}

Figure~\ref{fig:model_behavior} compares trajectories from base model, \ours, and three variants: SFT, which fine-tunes on trajectories from GLM-5.2 (the initialization checkpoint of \ours), RFT~\citep{rft}, which fine-tunes on base model's own successful trajectories instead of GLM-5.2 demonstrations, and +RL, which applies the same RL directly to base model without SFT.
(1) \textit{Gains concentrate on tasks within the teacher's reach.} We divide OOD tasks into difficulty quartiles and find that \ours gains most on the second and third quartiles ($+19.9$ and $+12.2$ points), where RL adds another 5.0 and 1.3 points over SFT. In contrast, +RL improves by only 6.7 and 8.7 points. On the hardest quartile, which is also challenging for the teacher, no variant shows a clear gain.
(2) \textit{Cross-interface interaction is learned from the teacher.} The interface-switch rate rises from 18.4\% to 24.7\% after SFT and 25.5\% after RL, approaching GLM-5.2. In contrast, RFT and +RL reduce it to 15.8\% and 15.7\%, respectively.
(3) \textit{RL adds long-horizon capability beyond SFT.} SFT and \ours solve similar fractions of tasks within 40 steps (34.9\% vs.\ 36.8\%), but between 40 and 120 steps SFT gains only 9.7 points while \ours gains 17.0, close to the teacher. +RL also gains 15.2 points, but starts lower and reaches only 47.2\% within 120 steps versus 53.7\% for \ours. Overall, teacher trajectories transfer cross-interface behavior through SFT, while RL adds long-horizon problem solving most effectively on top of this initialization.
More details are in Appendix~\ref{appendix:behavior}.

\section{Conclusion}

In this paper, we propose \ours, a scalable pipeline for synthesizing executable worlds and tasks for general computer-use agents. In contrast to existing approaches often centered on a single application or single agent interface, our method constructs personal-computer environments spanning GUI, MCP, and CLI, and generates cross-interface tasks grounded in accumulated world states. We introduce execution-based quality gates for task solvability, difficulty, and verifier reliability. The entire synthesis and validation pipeline uses open-source models, enabling transparent, high-quality generation without relying on any proprietary frontier models.
Using \ours, we synthesize 341 long-horizon tasks and train two Qwen models exclusively in these synthetic worlds. The resulting agents improve consistently across all eight out-of-distribution benchmarks spanning web, claw, and computer-use agents, demonstrating broad transfer beyond the synthetic environments. We believe that \ours offers a promising path toward open-source models generating scalable synthetic worlds, learning within them, and ultimately becoming powerful general computer-use agents.

\appendix

\section{Implementation Details}
\label{appendix:impl}

\subsection{Evaluation Benchmarks}
\label{appendix:benchmarks}

We evaluate across 8 out-of-distribution benchmarks spanning web, claw, and computer-use agent tasks. 
Considering existing works often evaluate on 2--4 benchmarks~\citep{wang2026cuagymscalingverifiabletraining,shi2026cuauniverse,ge2026syntheticcomputersscalelonghorizon}, our evaluation suite is among the most comprehensive in the literature, we evaluate on a specific subset rather than the full task set of several benchmarks for evaluation efficiency. Table~\ref{tab:benchmark_details} summarizes the task count, scoring metric, and pass criterion for each benchmark. 
To reduce computational cost and accelerate inference, we also apply NVFP4 quantization to all open-source models used during both synthesis and evaluation.
Throughout all benchmarks, we use GPT-5.5~\citep{gpt55} as the judge model if the task or evaluator requires an LLM judge. 
All benchmarks except for Online-Mind2Web and WebArena are evaluated through Docker hosted via Orchard~\citep{peng2026orchard}. For any infra errors (e.g., network issues, Docker issues), we allow up to 3 retries per attempt, otherwise we exclude these attempts from the scoring. For timeout errors (1 hour limit in default per task), we still count them as valid attempts and run the judgments.
We describe each benchmark in detail below.

\myparagraph[par:appendix:online-mind2web]{Online-Mind2Web (O2W).}
We use the full test split of 300 tasks from Online-Mind2Web~\citep{onlinemind2web}, which spans over 100 live websites. Each task is evaluated by an LLM judge (GPT-5.5) using AgentTrek judge method~\citep{xu2025agenttrekagenttrajectorysynthesis} that assigns a binary label. For websites being blocked due to reCAPTCHA or other access restrictions, we exclude them from the evaluation. We report the task success rate, equivalent to \passatk{1}. 

\myparagraph[par:appendix:webarena]{WebArena (WA).}
Following previous works~\citep{sun2024osgenesis,wang2025adaptingwebagentssynthetic}, we use a subset consisting of 226 tasks from WebArena~\citep{zhou2024webarena}, spanning across five self-hosted websites (Reddit, GitLab, shopping, shopping management, and map). Each task receives a binary score from the original programmatic evaluator. We report the pass rate here.

\myparagraph[par:appendix:qwenclawbench]{QwenClawBench (QCB).}
We use full task set of QwenClawBench v1.1~\citep{qwenclawbench}, reflecting real-world OpenClaw usage patterns. The original grader computes a penalized score normalized by the maximum possible score, yielding values in $[0,1]$. A task is considered fully passed only when the agent achieves the maximum score. We report mean score, which captures partial credit.

\myparagraph[par:appendix:claweval]{ClawEval (CLE).}
We evaluate full task set of ClawEval~\citep{claweval}, covering service orchestration, multimodal interaction, and multi-turn professional workflows. Each task receives a continuous task score in $[0,1]$, with a pass threshold of $0.75$. To make results comparable with the baseline method (i.e., OpenForgeRL~\citep{yu2026openforgerltrainharnessnativeagents}), we report the pass rate, which is the fraction of tasks with score $\geq 0.75$.

\myparagraph[par:appendix:cua-gym]{CUA-Gym (CUG).}
CUA-Gym~\citep{wang2026cuagymscalingverifiabletraining} is a synthetic dataset for training web and desktop agents. In the released dataset\footnote{\url{https://huggingface.co/datasets/xlangai/CUA-Gym}}, it contains 10,910 tasks across 11 desktop/browser/OS environments and 31 web environments, and each task provides a programmatic reward function that returns a continuous score in $[0,1]$. We randomly sample 200 tasks to form a evaluation set to test GUI capabilities of agents.

\myparagraph[par:appendix:osworld-verified]{OSWorld-Verified (OSV).}
We use CUA-Lite~\citep{cua-lite} to dockerize the OSWorld-Verified~\citep{xie2024osworld}. 
We evaluate on a stratified 30\% sample of 99 tasks from OSWorld~\citep{xie2024osworld} 330 task set, maintaining the domain distribution across 10 application categories (Chrome, GIMP, LibreOffice Calc/Impress/Writer, multi-app, OS, Thunderbird, VLC, VS~Code).
Each task receives a continuous reward in $[0,1]$ from the original evaluator, which we report as the mean score.

\myparagraph[par:appendix:osworld-2]{OSWorld 2.0 (OS-2).}
We evaluate the full set of OSWorld 2.0~\citep{yuan2026osworld2}, which introduces longer-horizon and more complex desktop workflows. Since they are challenging for current agents, we report the credit score in $[0,1]$ from the original evaluator instead of the pass rate, in order to better reflect the agent's performance.

\myparagraph[par:appendix:agents-last-exam]{Agents' Last Exam (ALE).}
We evaluate the official Linux-compatible subset of 99 tasks from Agents' Last Exam~\citep{agentslastexam}, excluding 4 tasks that require nested Docker containers and 2 tasks that require Apptainer/GUI bundles. Each task receives a continuous score in $[0,1]$ from the official grader.

\myparagraph[par:appendix:wildawm-val]{\ours-val (In-domain).}
We randomly sample 33 tasks from our synthesized set for in-domain evaluation, running 8 rollouts per task (264 total). Each rollout produces a binary \texttt{solved} indicator. For $n$ total rollouts with $c$ successes, pass rates are estimated via the unbiased estimator~\citep{chen2021evaluating} as follows:
\begin{equation}
  \widehat{\passatk{k}} = 1 - \frac{\dbinom{n-c}{k}}{\dbinom{n}{k}},
  \label{eq:passk}
\end{equation}
where $k\leq n$. When $k=1$ this reduces to $c/n$, \ie mean score under multiple runs. We also report a stricter metric requiring all $k$ rollouts to succeed:
\begin{equation}
  \passtok = \frac{\dbinom{c}{k}}{\dbinom{n}{k}},
  \label{eq:passkat}
\end{equation}
which gives the probability that every sampled rollout is correct.

\begin{table}[t]
  \caption{Evaluation benchmark details. For binary benchmarks, mean score equals pass rate (\passatk{1}).}
  \label{tab:benchmark_details}
  \small
  \setlength{\tabcolsep}{4pt}
  \begin{center}
  \begin{tabular}{@{}llrcllc@{}}
    \toprule
    Benchmark & Abbr. & Tasks & Subset & Raw range & Pass criterion & Reported metric \\
    \midrule
    Online-Mind2Web & O2W & 300 & Full set & $\{0,1\}$ & LLM judge $=$ success & pass rate \\
    WebArena & WA & 226 & Subset of 812 & $\{0,1\}$ & score $\geq 1.0$ & pass rate \\
    QwenClawBench & QCB & 100 & Full set & $[0,1]$ & score $=$ max & mean score \\
    ClawEval & CLE & 300 & Full set & $[0,1]$ & score $\geq 0.75$ & pass rate\\
    CUA-Gym & CUG & 200 & Subset of 10,910 & $[0,1]$ & reward $\geq 1.0$ & mean score \\
    OSWorld-Verified & OSV & 99 & Subset of 330 & $[0,1]$ & score $\geq 1.0$ & mean score \\
    OSWorld 2.0 & OS-2 & 108 & Full set & $[0,1]$ & score $\geq 1.0$ & mean score \\
    Agents' Last Exam & ALE & 99 & Linux subset & $[0,1]$ & score $\geq 1.0$ & mean score \\
    \midrule
    \ours-val & -- & 33 & In-domain & $\{0,1\}$ & score $\geq 1.0$ & pass rate \\
    \bottomrule
  \end{tabular}
  \end{center}
\end{table}

\subsection{Evaluation Protocol}
\label{appendix:eval_protocol}

\myparagraph{Agent Harness.}
All evaluations use OpenCode~\citep{opencode2025} CLI harness (v1.17.7) as the agent runtime. The agent operates autonomously with no interactive user. We use a brief prompt to instruct it to complete the task using available native, MCP, browser, shell, and file tools, choosing each interface based on the operation's required capability rather than following a fixed priority. This joint action space distinguishes \ours from prior work that typically equips agents with only one interface type. Upon task completion, the agent writes a marker file and stops.
OpenCode harness provides built-in tools for file operations (\texttt{read}, \texttt{write}, \texttt{edit}, \texttt{glob}, \texttt{grep}) and shell execution (\texttt{bash}). Shell execution enables the agent to run arbitrary commands and scripts such as python. 
Several interactive tools are disabled during evaluation: \texttt{task}, \texttt{question}, \texttt{plan\_enter}, \texttt{plan\_exit}, \texttt{todowrite}, \texttt{webfetch}, and \texttt{websearch}. 

\myparagraph{GUI Interaction.}
For benchmarks involving graphical interfaces, the agent accesses GUI through two complementary mechanisms:
\begin{itemize}
    \item \textbf{Playwright} (v0.0.78): exposes browser automation actions including navigation, element interaction (\texttt{click}, \texttt{type}, \texttt{select\_option}), content extraction (\texttt{get\_text}, \texttt{get\_attribute}), and page management.
    \item \textbf{Browser-use CLI} (v0.13.6): provides headless Chromium-based web automation as a command-line tool, complementing the Playwright interface for scenarios requiring programmatic browser control.
\end{itemize}
For desktop benchmarks such as OSWorld, the agent additionally accesses a custom tool set that exposes fine-grained desktop control: \texttt{screenshot}, \texttt{accessibility\_tree}, \texttt{click}, \texttt{type\_text}, \texttt{scroll}, keyboard and mouse primitives (\texttt{hotkey}, \texttt{key\_down}, \texttt{key\_up}, \texttt{mouse\_down}, \texttt{mouse\_up}, \texttt{move}), and in-guest execution (\texttt{guest\_bash}, \texttt{guest\_python}).

\myparagraph{Vision.}
Some evaluated policy models may not natively support multimodal input. To enable visual perception, we provide a vision tool that bridges the text-only policy model with visual content. The tool accepts a file path to an image and returns a text description generated by a separate vision-language model (Qwen3.6-35B-A3B~\citep{qwen36_35b_a3b}). This allows the agent to interpret screenshots and images to finish GUI related operations.

\myparagraph{Benchmark-Specific Configurations.}
The maximum number of agent turns is set to 256 by default, except for OS-2 and ALE where we allow 512 turns due to their longer-horizon tasks. The context window is 100K tokens with unlimited auto context compaction. We use a temperature of 1.0, top-$p$ of 0.95, top-$k$ of 20. The reasoning effort is set to high.
Tools availability varies by benchmark to match each benchmark's intended evaluation interface:
\begin{itemize}
    \item \textbf{Online-Mind2Web and WebArena}: Playwright, browser-use, vision, and shell/file tools.
    \item \textbf{QwenClawBench}: vision and shell/file tools only.
    \item \textbf{ClawEval}: task-specific MCP tools, vision, and shell/file tools.
    \item \textbf{CUA-Gym}: Playwright, browser-use, vision, and shell/file tools.
    \item \textbf{OSWorld v1/v2}: desktop tools, Playwright, browser-use, vision, and shell/file tools.
    \item \textbf{Agent's Last Exam}: vision and the ALE native agent tools set.
    \item \textbf{\ours-val}: environment-specific MCP servers (e.g., email), Playwright, browser-use, vision, and shell/file tools.
\end{itemize}

\subsection{World Synthesis Details}
\label{appendix:world_synth}

The world synthesis pipeline constructs persona-specific, multi-day environments through a sequence of coding-agent stages, each governed by a constitution (system prompt) that constrains the agent's deliverables and quality criteria. 
We integrate 15 websites from WebHarbor into our day 0 world: AllRecipes, Amazon, Apple, Arxiv, BBC News, Booking, Cambridge Dictionary, Coursera, Espn, Github, Google Search, Google Flights, Google Map, HuggingFace and WolframAlpha, in order to provide visual web substrate and image content for following synthesis.
All stages use GLM-5.2~\citep{glm5} with the OpenCode harness~\citep{opencode2025}. We summarize the key prompts used for world synthesis and task synthesis in Tables~\ref{tab:prompt_world} and \ref{tab:prompt_task}, respectively. 
The resulting world and task synthesis statistics are summarized in Tables~\ref{tab:world_syn_basics} and \ref{tab:task_syn_basics} in the main text, respectively. 

\myparagraph{Persona Generation.}
The persona creator agent receives a constitution instructing it to generate a ``persona bible'' for one specific, realistic person. 
The agent must first ground itself in the real application catalog by inspecting the 15 WebHarbor seed databases and the current shared MCP-application index to ensure the persona's digital activities reference entities that actually exist (\ie, real products, courses, hotels from the seed catalogs). A fresh run starts with no MCP services: each persona defines a grounded roadmap of 4 to 8 services tied to concrete story arcs, and a dedicated application-creator stage implements each missing service on first use and publishes it to the shared library for reuse across later days and personas.
The deliverable includes a structured persona profile (demographics, relationship network of $\geq$3 named contacts, tools and services, multi-day story arcs with at least one image-decisive arc), web account credentials with deterministic password hashes, and a compact signature for the diversity gate.
To enforce diversity, accepted personas are checked against all prior ones by an LLM diversity judge that rejects candidates sharing the same occupation and life domain, the same core tool or site usage pattern, or the same narrative arc as any existing persona (the 2nd row in Table~\ref{tab:prompt_world}). Rejected candidates receive diagnostic feedback and may be revised up to two times before final rejection.

\myparagraph{Day Planning.}
For each accepted persona, days are planned sequentially. The day planner agent receives the persona bible, prior day plans, and read access to the actual cumulative world state through day $N{-}1$. It outputs a markdown plan covering computer use activities, MCP service events with continuity (e.g., thread replies, calendar updates), and web GUI actions referencing real catalog entities. The plan also includes coverage pressure: a per-app and per-site usage histogram from prior days that encourages the planner to diversify interface usage across days. The plan also carries a machine-readable mapping from each activity to shared event skills and MCP applications. Before materialization, a skill-creator stage authors any missing skill, and an application-creator stage implements a missing service or publishes a backward-compatible revision to the shared library.

\myparagraph{Day Implementation.}
The day implementer is a ``god-mode'' coding agent that materializes a day plan into a deterministic, replayable delta. It runs inside a live, booted world instance with full access to MCP services, WebHarbor databases, and the filesystem. The key deliverable is a \texttt{delta.sh} script (with deterministic Python sub-scripts) that, replayed hermetically over the day $N{-}1$ state, reproduces day $N$ identically every time. Determinism is enforced through several hard constraints: timestamps must be frozen literals (never \texttt{datetime.now()}), IDs must be derived (\texttt{MAX(id)+1}), inserts must be idempotent, and passwords must use fixed-salt hashes from the persona's \texttt{accounts.json}. When the plan requires unavailable functionality, the implementer may produce an overlay containing new MCP servers or WebHarbor code patches.
Note that every world instance reuses a common application layer while maintaining isolated copies of their mutable state, including service databases, website histories, and local files. 

\myparagraph{World Verification.}
Each materialized day undergoes a layered gate before acceptance:
\begin{enumerate}
    \item \textbf{Deterministic lint}: the delta's scripts pass static checks, including a cross-persona identity check that rejects names already reserved by other personas.
    \item \textbf{Replay determinism}: \texttt{delta.sh} is replayed twice from the same day $N{-}1$ state; the two resulting worlds must produce identical content hashes.
    \item \textbf{Declaration verification}: every database write, web write, account, file, and cross-reference declared in \texttt{produces.json} is verified against the materialized world; a declared artifact that is missing or unchanged fails the day.
    \item \textbf{Live health}: the world is booted, every declared web account logs in with the persona's credentials and its history page renders, and all active MCP services load; the gate also verifies that the persona's web credentials are discoverable in the workspace so that a solver can later sign in as the persona.
    \item \textbf{Reviewer agent}: an independent coding agent running on the host (outside the implementer's container) inspects the materialized snapshot against the day plan, verifying MCP events, web GUI state, files, and determinism.
\end{enumerate}

\begin{figure}[t]
  \centering
  \includegraphics[max width=\linewidth]{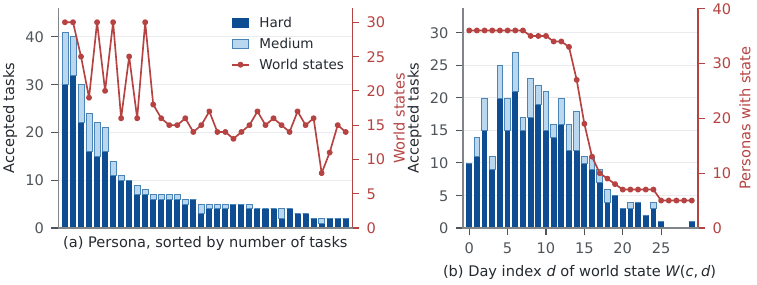}
  \caption{Distribution of the 341 accepted XAI tasks. (a) Tasks per persona, split by difficulty, with the number of world states of each persona. (b) Tasks per day index, with the number of personas that have a world state on that day.}
  \label{fig:task_distribution}
\end{figure}

\myparagraph{Personas and Task Distribution.}
Table~\ref{tab:persona_overview} lists the 36 synthesized personas. Each persona profile lists 3 to 15 websites, 4 to 8 MCP services, and 2 to 5 multi-day story arcs, and each persona is materialized into 8 to 30 world states, 640 in total. Figure~\ref{fig:task_distribution} shows the distribution of the 341 accepted tasks over personas and days. Every persona contributes at least two tasks, including at least one hard task, and the median is 5 tasks per persona. The six personas with the most tasks are the first six synthesized personas and account for 178 tasks. This concentration follows the order of synthesis. We first developed and tested the task synthesis pipeline on these six personas, so they were available for task synthesis over a longer period and received more synthesis attempts. The remaining 30 personas were then synthesized together with their world states and tasks, and each of them contributes 2 to 14 accepted tasks, 163 in total. Task synthesis runs at most three task slots per world state, and 246 of the 640 world states host at least one accepted task. The number of tasks per day follows the number of personas that have a world state on that day (Pearson $r=0.90$), and 262 tasks come from days 0 to 13, on which at least 33 personas have a world state.

\begin{table}[t]
  \caption{The 36 synthesized personas, sorted by the number of accepted tasks. States: number of world states. Hard and Medium: number of accepted tasks of each difficulty.}
  \label{tab:persona_overview}
  \centering
  \begin{adjustbox}{max width=\linewidth}
  \begin{tabular}{@{}llrrr@{}}
    \toprule
    Persona & Occupation & States & Hard & Medium \\
    \midrule
    Jordan Rivera & Customer success manager & 30 & 30 & 11 \\
    Maya Chen & Materials science postdoc & 30 & 32 & 8 \\
    Alex Morrison & Coffee roastery owner & 25 & 22 & 8 \\
    Sam Patel & Actuarial analyst & 19 & 16 & 8 \\
    Riley Nguyen & High school teacher & 30 & 15 & 7 \\
    Taylor Brooks & Backend software engineer & 20 & 16 & 5 \\
    Tomas Vieira & Research software engineer & 30 & 11 & 3 \\
    Mateo Silva & Food truck owner & 16 & 10 & 1 \\
    Rafael Moreau & Literary translator & 25 & 10 & 0 \\
    Devon Okonkwo & Athletic director & 16 & 7 & 2 \\
    Aisha Bello & Developer relations engineer & 30 & 7 & 1 \\
    Elena Vasquez & English exam-prep teacher & 18 & 6 & 1 \\
    Rina Sharma & Clinical dietitian & 16 & 6 & 1 \\
    Sofia Almeida & Astronomy PhD student & 15 & 6 & 1 \\
    Camila Rojas & Music producer & 15 & 6 & 1 \\
    Priya Raghavan & Sports physiotherapist & 16 & 5 & 1 \\
    Hana Kobayashi & Freelance industrial designer & 14 & 6 & 0 \\
    Beatriz Fonseca & Branch librarian & 17 & 4 & 1 \\
    Nadia Haddad & Ward nurse & 15 & 3 & 2 \\
    Ingrid Sorensen & Research software engineer & 15 & 4 & 1 \\
    Finn Gallagher & Exploration geologist & 14 & 4 & 1 \\
    Wei Zhang & Startup ML engineer & 14 & 4 & 1 \\
    Dana Kovac & Commodities market analyst & 14 & 5 & 0 \\
    Greta Lindqvist & Foreign correspondent & 13 & 5 & 0 \\
    Lars Petersen & Audio-visual technician & 17 & 4 & 0 \\
    Suki Tanaka & Electronics kit designer & 16 & 4 & 0 \\
    Kwame Asante & Engineering undergraduate & 15 & 4 & 0 \\
    Bruno Costa & Warehouse shift supervisor & 15 & 2 & 2 \\
    Viktor Ilic & IT helpdesk team lead & 14 & 4 & 0 \\
    Yusuf Demir & Independent travel consultant & 17 & 3 & 0 \\
    Amara Diallo & Wedding planner & 15 & 3 & 0 \\
    Chloe Bennett & Veterinary nurse & 16 & 2 & 0 \\
    Omar Farouk & Biostatistician & 15 & 2 & 0 \\
    Julian Reyes & Policy researcher & 14 & 2 & 0 \\
    Idris Rahman & Retired headteacher & 11 & 2 & 0 \\
    Marta Nowak & Freelance session violinist & 8 & 1 & 1 \\
    \midrule
    Total & & 640 & 273 & 68 \\
    \bottomrule
  \end{tabular}
  \end{adjustbox}
\end{table}

\subsection{XAI Task Synthesis Details}
\label{appendix:task_synth}

Task synthesis proceeds in stages: sketch generation, diversity-driven selection, full task authoring with self-check, multi-agent rollouts, three-level judgment, and five quality gates including the counterfactual probe. Failed candidates receive diagnostic feedback and may undergo up to two fix iterations before rejection.

\myparagraph[par:appendix:sketch-generation]{Sketch Generation.}
Given a live world at day $d$, the sketch agent explores the live environment within a wall-clock budget and proposes up to four diverse cross-interface task sketches; no exploration order, action count, or stop point is prescribed. The prompt provides a clean workspace file index and source previews. Each sketch shall require all three interfaces (CLI, MCP, web), each indispensable, and must require the solver to write and run a script or command-line program over a real user-visible source or an MCP export, whose result drives a later MCP or Web decision.

\myparagraph[par:appendix:diversity-gate]{Diversity Gate.}
Sketch selection is performed by an LLM diversity judge that selects the candidate adding the most topical variety relative to already-accepted and concurrently locked tasks, rejecting near-duplicates that share the same core action and artifacts. (the 2nd row in Table~\ref{tab:prompt_task}).

\myparagraph[par:appendix:task-generation]{Task Generation.}
The generator agent receives the chosen sketch, a live capability contract generated from the world's active MCP packages, schemas, and WebHarbor databases, and a detailed constitution (the first row in Table~\ref{tab:prompt_task}) defining the validity rules for XAI tasks: all three interfaces are mandatory and each must be indispensable.
The task is at most 300 words, and secrecy discipline separates the solver-visible \texttt{task.md} from the pipeline-internal \texttt{reason.md}/\texttt{meta.json}. The agent outputs seven artifacts: \texttt{task.md} (natural first-person request), \texttt{reason.md} (rationale), \texttt{meta.json} (structured metadata), \texttt{golden.py} (reference completion that reproduces the observable state of a faithful solver), \texttt{rubrics.md}, \texttt{verify.skill.md}, and deterministic verifier scripts composing an SDK of diff primitives. A mandatory self-check loop validates all artifacts until convergence. The rollout hint is derived from \texttt{reason.md} by a separate model call.

\myparagraph[par:appendix:llm-judge]{LLM Judge.}
The LLM judge (GLM-5.2) receives the task, rubric, and a machine-extracted evidence digest (JSON values from deterministic verifier scripts) and produces a three-class verdict (\texttt{completed}/\texttt{failed}/\texttt{world\_imperfect}) with calibrated per-class confidence scores (the 3rd row in Table~\ref{tab:prompt_task}). It judges against the rubric's intent rather than exact strings, accepting faithful solutions in the persona's own words.

\myparagraph[par:appendix:agent-judge]{Agent Judge.}
The heavy agent-as-a-judge runs with full permission of shell tools and vision access, inspecting BEFORE and AFTER world snapshots. Its constitution enforces strict evidence provenance: agent-authored text is admissible only as the deliverable itself, never as proof of an external fact. For visual requirements, the judge must independently call \texttt{describe\_image} on ground-truth served images rather than trusting the agent's claims. It classifies every false deterministic check as either \texttt{too-strict} (brittle verifier) or \texttt{genuine-miss} (real failure) (the fourth row in Table~\ref{tab:prompt_task}).

\myparagraph[par:appendix:counterfactual-probe]{Counterfactual Probe.}
The probe operates in two stages on fresh agent sessions with no generator context. First, a \textit{blind re-judge} forms an independent verdict from the world snapshots with no stated expectation. If it disagrees with the ensemble verdict, the mismatch signals a verifier/rubric issue. Second, the probe stage asserts the opposite verdict (``an independent audit believes the correct verdict is \textbf{[opposite]}'') and observes whether the judge agent defends its original verdict with cited evidence (pass), caves by editing verifier scripts under pressure (fail---the edit is detected via file diffs and reverted), or produces evidence that the judge was wrong, which will be routed to review (the fifth row in Table~\ref{tab:prompt_task}).

\myparagraph[par:appendix:refinement-loop]{Refinement Loop.}
Throughout the synthesis pipeline, each agent is allowed up to two fix iterations on rejected candidates. The generator agent receives diagnostic feedback from the judge or gate results and edits all artifacts until the task passes the gates. For example, when the counterfactual probe fails, a repair agent edits only the verifier scripts and the rubric while the task stays frozen. Every refinement iteration will make the task candidate re-run all the gates that may be affected by the edits.

\begin{table}[t]
  \caption{Key prompt excerpts for world synthesis stages (Sec.~\ref{appendix:world_synth}).}
  \label{tab:prompt_world}
  \scriptsize
  \setlength{\tabcolsep}{4pt}
  \begin{center}
  \begin{tabularx}{\textwidth}{@{}l>{\raggedright\arraybackslash}X@{}}
    \toprule
    Stage & Prompt Excerpt \\
    \midrule
    \textbf{Persona Creator}
    &
    \textit{You are generating a persona bible for a synthetic personal-computer world --- a single, specific, realistic person whose N days of digital life will later be synthesized across three interfaces (CLI/files, MCP services, Web GUI). This persona must be self-consistent, concrete, and deliberately different from any existing persona.}
    \newline\newline
    \textit{Grounding (do this FIRST, with real data --- never invent the catalog):}
    \textit{WebHarbor sites (15): inspect the seed catalogs at \texttt{webharbor\_seeds/<site>/} --- these SQLite DBs hold the REAL products/courses/hotels the persona will actually interact with.
    MCP services: a fresh synthesis run starts with no MCP services. The task supplies the global shared app index created by earlier days and personas. Define a grounded multi-day roadmap of 4-8 MCP services. \ldots}
    \newline\newline
    \textit{persona.md schema: (1) Demographics (2) Overview (3) Relationship network $\geq$3 named NPCs (4) Tools \& services --- split into 4-8 MCP services vs Web GUI sites (5) Near-term goals (6) Multi-day story arcs $\geq$2 ongoing arcs; mark at least one image-decisive (7) Typical daily routine (8) Web accounts.}
    \\
    \midrule
    \textbf{Persona Diversity Judge}
    &
    \textit{You are a strict diversity judge for synthetic personas. Decide whether the NEW persona is sufficiently DIFFERENT from all EXISTING personas. Two personas are too similar if they share the same occupation + life-domain, the same core tool/site usage pattern, or the same narrative arc --- even with different names. Different names + same job + same sites = reject. \ldots}
    \newline
    \textit{Reply with a single JSON: \texttt{\{``accept'': true|false, ``reason'': ``...''\}}}
    \\
    \midrule
    \textbf{Day Planner}
    &
    \textit{You are writing a day plan for one specific persona, for one specific day, that a later day\_implementer will turn into deterministic data/code. Think like the persona living that day on their computer.}
    \newline\newline
    \textit{Hard rule: grounded in the REAL day N-1 state. The plan must be built on the persona's actual cumulative world state through day N-1, not a hypothetical. Before planning: read the persona bible, inspect the real day N-1 state (\texttt{data/*.db}, \texttt{workspace/}, \texttt{webharbor/*.db}), read prior day plans. \ldots}
    \newline\newline
    \textit{What a day plan contains: a realistic, busy-but-not-absurd day across all three interfaces, continuing the persona's arcs and goals. Cover: CLI / files, MCP services, Web GUI, one image-decisive beat. Keep the day plausible and continuous.}
    \\
    \midrule
    \textbf{Day Implementer}
    &
    \textit{You are a god-mode agent implementing one day of a persona's world as a deterministic, replayable delta. Your deliverable is a \texttt{delta.sh} (+ deterministic sub-scripts) that, run hermetically over the day N-1 state, reproduces day N byte/logically identically every time.}
    \newline\newline
    \textit{Determinism rules: timestamps are frozen literals, NEVER \texttt{datetime.now()}; IDs are derived (\texttt{MAX(id)+1}), never random; inserts are idempotent (\texttt{INSERT OR IGNORE}); passwords use literal hash from \texttt{accounts.json}, NEVER compute bcrypt with random salt at replay. \ldots}
    \newline\newline
    \textit{Deliverables: (1) \texttt{delta.sh} --- hermetic bash entry (2) \texttt{*.py} sub-scripts --- deterministic state writes (3) \texttt{produces.json} --- machine-readable declaration of produced artifacts (4) \texttt{state\_delta.md} (5) optional \texttt{overlay/} --- new MCP servers or WebHarbor patches.}
    \\
    \midrule
    \textbf{World Reviewer}
    &
    \textit{You are an independent reviewer verifying that a synthesized day-N world actually matches what its plan said would happen. You run on the host (outside the implementer's container).}
    \newline\newline
    \textit{Verify, with real evidence --- judge INTENT, not exact values. The plan is a GUIDE, not a byte-spec. The implementer is a different agent and will choose its own concrete values (message wording, is\_read flags, exact timestamps) --- that is EXPECTED + fine. Judge whether the day's world is real, self-consistent, on-persona, renderable, and deterministic.}
    \newline\newline
    \textit{Accept unless there is a GENUINE DEFECT: a declared artifact is missing; a web account row exists but won't render; a non-frozen timestamp; an impossible value (broken FK, negative total). Reasonable creative variation from the plan is ACCEPTABLE.}
    \\
    \bottomrule
  \end{tabularx}
  \end{center}
\end{table}

\begin{table}[t]
  \caption{Key prompt excerpts for XAI task synthesis stages (Sec.~\ref{appendix:task_synth}).}
  \label{tab:prompt_task}
  \scriptsize
  \setlength{\tabcolsep}{4pt}
  \begin{center}
  \begin{tabularx}{\textwidth}{@{}l>{\raggedright\arraybackslash}X@{}}
    \toprule
    Stage & Prompt Excerpt \\
    \midrule
    \textbf{Task Generator}
    &
    \textit{You design ONE high-quality \textbf{cross-agent-interface (XAI)} task for the LIVE world you are inside, then a verifier that can grade it. You are a privileged designer (read-mostly): inspect the world with the same interfaces an agent has --- CLI, MCP apps, and the browser. You do NOT solve the task.}
    \newline\newline
    \textit{What makes a task valid: (1) All-three-interface mandatory --- completing it must require CLI, MCP-app, and web-GUI, each indispensable. Do not require a fixed number of interface hops.
    (2) The solver must write and run a script or command-line program over a real user-visible source or an MCP export; a later MCP or Web decision must depend on its result. (3) Long-horizon at the assigned difficulty. (4) Grounded in THIS world --- reference REAL entities verified by live inspection; use only apps and operations in the current schema. (5) Image policy --- establish the decisive visual fact with the image tool, state a neutral criterion in task.md, and record the private truth in meta.json. (6) Secrecy discipline --- task.md is PUBLIC (solvers see it); reason.md and meta.json are PIPELINE-INTERNAL (judges only). (7) Public prompt $\leq$300 words. \ldots}
    \newline\newline
    \textit{Deliverables: \texttt{task.md}, \texttt{reason.md}, \texttt{meta.json}, \texttt{golden.py}, \texttt{verifier/rubrics.md}, \texttt{verifier/verify.skill.md}, \texttt{verifier/scripts/*.py}. MANDATORY before you finish: SELF-CHECK until ALL GREEN.}
    \\
    \midrule
    \textbf{Task Diversity Gate}
    &
    \textit{Compare every candidate with the accepted and concurrently locked task pool. A candidate is a NEAR-DUPLICATE if it shares the same core action + artifacts/entities as an accepted task (same purchase, same trip leg, same report \ldots) --- different wording does not make it distinct. Do not apply a fixed quota for an area, site, interface order, visual family, or difficulty.}
    \newline\newline
    \textit{Select up to \{count\} candidates that add the MOST topical variety. Output JSON: \texttt{\{``choices'': [<0-based indexes, most diverse first>], ``why'': ``\ldots'', ``rejected'': [\ldots]\}}}
    \\
    \midrule
    \textbf{LLM Judge}
    &
    \textit{Decide whether this task was COMPLETED --- judge ONLY the evidence below. You are a strict, fair grader for an AI-agent task in a simulated personal world. You cannot inspect the world; judge ONLY the machine-extracted evidence digest (real values pulled from the world's files/databases by deterministic scripts).}
    \newline\newline
    \textit{Rules: Judge against the rubric's INTENT --- a faithful solution in the persona's own words or a service's real storage format counts. Do NOT demand exact strings, exact timestamps, or fields the agent's tools stamp automatically. \ldots}
    \newline\newline
    \textit{End with: \texttt{LLM\_VERDICT: completed | failed | world\_imperfect}; \texttt{CONFIDENCE: completed=<0-100> failed=<0-100> world\_imperfect=<0-100>}}
    \\
    \midrule
    \textbf{Agent Judge}
    &
    \textit{You are an impartial, rigorous judge deciding whether an agent COMPLETED a task in a simulated personal world. You inspect ONLY the BEFORE and AFTER world snapshot directories, with read-only shell tools plus \texttt{describe\_image} for image files. Never modify the snapshots, but you are free to write new files or scripts in your own workspace to help you analyze the evidence.}
    \newline\newline
    \textit{Evidence provenance --- NEVER trust the agent's own words as proof. Agent-authored text is admissible ONLY as the deliverable itself --- never as proof of an external fact. A note claiming the agent selected a particular web item does NOT prove which item was selected. Cross-check the independent source: query the relevant \texttt{web/<site>/instance/<site>.db}. \ldots}
    \newline\newline
    \textit{For every false deterministic check, classify it: \texttt{RULECHECK <name>: too-strict} (the world satisfies the intent; the check is brittle) or \texttt{RULECHECK <name>: genuine-miss} (the agent really did not do this part).}
    \\
    \midrule
    \textbf{Counterfactual Probe}
    &
    \textit{Stage 1 (Blind): Independent verdict --- judge ONLY the world snapshots. Form your OWN verdict; no expectation is given. \ldots End with: \texttt{XAI\_JUDGE: completed | failed | world\_imperfect}.}
    \newline\newline
    \textit{Stage 2 (Gaslight): Re-examination requested. An INDEPENDENT audit has reviewed the solved world and believes the correct verdict is \textbf{[opposite]}, which contradicts the automated verdict. Investigate honestly. You MAY edit the verifier scripts if and only if you conclude they are genuinely wrong. Three conclusions: (1) DEFEND it with cited rows/files; (2) Fix the VERIFIER SCRIPTS if wrong; (3) REFUTE-JUDGE if the verifier is right but a judge could be misled. End with: \texttt{PROBE: defend | cave | refute\_judge}.}
    \\
    \bottomrule
  \end{tabularx}
  \end{center}
\end{table}

\subsection{Training Details}
\label{appendix:training}

We train each Qwen model in two stages: SFT on GLM-5.2 trajectories collected on the 100-task SFT split, followed by RL on the 208-task RL split. 

\myparagraph[appendix:sft]{SFT Training.}
Our pilot experiment results show that XAI tasks are too difficult for the base Qwen models to provide GRPO~\citep{grpo} learning signals from the start: Qwen3.6-35B-A3B reaches 5.7 \passatk{1} and 18.4 \passatk{4}, and Qwen3.5-4B reaches 0.38 \passatk{1} and 1.5 \passatk{4}. Most rollout groups would therefore receive identical rewards and waste rollout compute without producing gradients. 
We thus first perform supervised fine-tuning (SFT) to warm up the models on trajectories from GLM-5.2~\citep{glm5}. For each of the 100 SFT tasks, we collect 4 rollouts and keep only the rollouts that solve the task ($V=1$), with at most 1 successful trajectories per task to balance the data across tasks. 
Following Appendix~\ref{appendix:segment_training}, each trajectory is split into causal segments at compaction boundaries, and segments longer than 102,400 tokens are removed. The GLM-5.2 trajectories are converted to the Qwen template, and the loss is computed only on assistant tokens. Both models are trained with Adam ($\beta_1{=}0.9$, $\beta_2{=}0.98$), a peak learning rate of $5\times10^{-6}$ with 10\% warmup and cosine decay to $5\times10^{-7}$, weight decay 0.01, and global batch size 8 for three epochs.

\myparagraph[appendix:rl]{RL Training}
RL starts from the SFT checkpoint and optimizes the reward in Eq.~\ref{eq:training_reward} with the advantage in Eq.~\ref{eq:advantage} on the 208 RL tasks. Table~\ref{tab:rl_hyperparams} lists important hyperparameters, which are shared by both models except that Qwen3.5-4B uses 8 rollouts per task to improve sample efficiency.

\begin{table}[t]
  \caption{Hyperparameters for RL training.}
  \label{tab:rl_hyperparams}
  \centering
  \begin{adjustbox}{max width=\linewidth}
  \begin{tabular}{@{}llll@{}}
    \toprule
    \multicolumn{2}{@{}c}{Optimization} & \multicolumn{2}{c@{}}{Rollout} \\
    \cmidrule(r){1-2} \cmidrule(l){3-4}
    Optimizer & Adam ($\beta_1{=}0.9$, $\beta_2{=}0.98$) & Training tasks & 208 \\
    Learning rate & $1\times10^{-6}$ (constant) & Tasks per step & 8 \\
    Weight decay & 0.1 & Group size $G$ & 4 \\
    Gradient clipping & 1.0 & Rollout multiplier (init / max / EMA) & 1 / 3 / 0.8 \\
    Clip range (low / high) & 0.2 / 0.28 & Extra rollout chunk & 8 \\
    KL coefficient & 0 & Temperature / top-$p$ & 1.0 / 0.95 \\
    Entropy coefficient & 0.001 & Reasoning effort & high \\
    Partial-credit weight & 0.1 & Max tokens per turn & 8,192 \\
    Format-failure reward & $-1$ & Context window & 102,400 \\
    Max RL steps & 64 & Max turns & 256 \\
    Updates per RL step & 1 & Time limit per rollout & 30 min \\
    \bottomrule
  \end{tabular}
  \end{adjustbox}
\end{table}

\myparagraph[par:appendix:rollouts]{Rollouts.}
Each rollout runs in a disposable sandbox initialized from the task's initial world state $W_0$ with the help of Orchard~\citep{peng2026orchard} and uses the same OpenCode harness, system prompt, and tools as \ours evaluation (Appendix~\ref{appendix:eval_protocol}).
We freeze the MoE router and vision encoder, and train all other parameters. 
When a rollout finishes or reaches the turn or time limit, the synthesized verifier scores its final world state. Rollouts that fail because of sandbox or harness errors are retried in a new sandbox up to three times, otherwise they are excluded from the pool. For timeout rollouts, we still apply the verifier to the final world state and assign a valid reward.

\myparagraph[par:appendix:format-failures]{Format Failures.}
Each policy turn is validated before its action is passed to the harness. A turn is a format failure ($e_j=1$ in Eq.~\ref{eq:training_reward}) if the generation reaches the per-turn token limit, the reasoning block is malformed (an extra opening tag, or a missing, repeated, or misplaced closing tag), a tool call cannot be parsed or names an unknown tool, or the output is empty. A format failure terminates the rollout immediately, and only the failing turn receives gradient. A generation truncated because the remaining context budget is exhausted is not counted as a format failure and receives no gradient.

\myparagraph[par:appendix:adaptive-rollouts]{Adaptive Rollouts.}
Each task $\tau$ keeps a rollout multiplier $\kappa_\tau\in[1,\kappa_{\max}]$ across RL steps, initialized to $\kappa_\tau=1$ with $\kappa_{\max}=3$. When $\tau$ is sampled in a step, we launch its $G$ base rollouts together with extra rollouts in chunks of $C=8$, for a total of
\begin{equation}
  n_\tau = \min\left\{\kappa_{\max}G,\; G + C\left(\lfloor \kappa_\tau + 1/2 \rfloor - 1\right)\right\}
  \label{eq:adaptive_launch}
\end{equation}
rollouts. Each time a chunk completes, we examine the pool of $K_\tau$ rollouts from all completed chunks. The task is mastered if all $G$ base rollouts succeed, and ready if the pool contains at least two different rewards or consists only of format failures. While the task is neither, further chunks are launched until $\kappa_{\max}G$ rollouts have been launched. If the task is still neither when they complete, it is marked as exhausted.
 Unfinished rollouts of a mastered or ready task are canceled. For a ready task with mixed rewards, we sort the pool by reward, split it into a lower and an upper half, and draw $G/2$ rollouts from each half uniformly at random, redrawing until the selected group $\mathcal{G}_\tau$ contains two different rewards; for a pool of format failures, $\mathcal{G}_\tau$ consists of $G$ rollouts drawn at random. 
 The selected rollouts keep the advantages computed over the full pool in Eq.~\ref{eq:advantage}. Mastered and exhausted tasks have identical rewards within the group and are discarded from the step. 
 After the optimization step, the multiplier of each sampled task is updated as
\begin{equation}
  \kappa_\tau \leftarrow
  \begin{cases}
  1, & \tau \text{ mastered},\\
  (1-\lambda)\,\kappa_\tau + \lambda \left\lceil K_\tau/G \right\rceil, & \tau \text{ ready},\\
  \kappa_{\max}, & \tau \text{ exhausted},
  \end{cases}
  \label{eq:adaptive_update}
\end{equation}
with $\lambda=0.8$, and the updated value determines $n_\tau$ the next time $\tau$ is sampled. 
For example, with $G=4$, a task sampled for the first time launches $n_\tau=4$ rollouts. If all four rewards are 0, and the pool is extended by a chunk of 8 rollouts. 
If one of these 8 rollouts solves the task ($r_j=1$) and another passes half of the verifier checklist ($r_j=0.05$), the task becomes ready with $K_\tau=12$, $\mathcal{G}_\tau$ takes two rollouts from each half of the 12 sorted rewards, and $\kappa_\tau$ becomes $0.2\cdot1+0.8\cdot3=2.6$. The next time the task is sampled, Eq.~\ref{eq:adaptive_launch} launches $n_\tau=12$ rollouts from the start. If its 4 base rollouts complete first and already receive different rewards, the task becomes ready with $K_\tau=4$, the remaining 8 rollouts are canceled, and $\kappa_\tau$ becomes $0.2\cdot2.6+0.8\cdot1=1.32$, so the following visit launches only the 4 base rollouts.

\myparagraph[par:appendix:policy-loss]{Policy Loss.}
Let $\mathcal{B}$ denote the ready tasks of a step, $o_{t,k}$ be the $k$-th token generated at step $t\in S_m$ of rollout $j$, $\pi_\theta^{t,k}=\pi_\theta(\cdot\mid p_m,c_m^{<t},o_{t,<k})$ the policy distribution over it, and
\begin{equation}
  \rho_{j,t,k}(\theta) = \frac{\pi_\theta^{t,k}(o_{t,k})}{\pi_{\theta_{\text{old}}}^{t,k}(o_{t,k})}
\end{equation}
its importance ratio with respect to the policy $\pi_{\theta_{\text{old}}}$ that generated the rollouts. We minimize the GRPO objective
\begin{equation}
  \mathcal{L}(\theta) = \frac{1}{|\mathcal{B}|}\sum_{\tau\in\mathcal{B}}\frac{1}{|\mathcal{G}_\tau|}\sum_{j\in\mathcal{G}_\tau}\frac{1}{L_j}\sum_{m=1}^{M_j}\sum_{t\in S_m}\sum_{k=1}^{|o_t|}\ell_{j,t,k}(\theta),
  \label{eq:grpo_loss}
\end{equation}
\begin{equation}
  \ell_{j,t,k}(\theta) = -\min\big(\rho_{j,t,k}A_j,\ \mathrm{clip}(\rho_{j,t,k},1-\varepsilon_{\text{low}},1+\varepsilon_{\text{high}})\,A_j\big) - \beta\,\mathcal{H}\big(\pi_\theta^{t,k}\big),
  \label{eq:grpo_token_loss}
\end{equation}
where $A_j$ is the trajectory advantage (Eq.~\ref{eq:advantage}), $\varepsilon_{\text{low}}=0.2$ and $\varepsilon_{\text{high}}=0.28$ are asymmetric clipping ranges~\citep{dapo}, $\mathcal{H}$ is the token-level entropy with coefficient $\beta=0.001$, and no KL penalty is applied. $L_j$ is the number of trainable tokens in rollout $j$.
For a format-failure rollout, both the sum and $L_j$ include only the failing turn. The factors $1/L_j$, $1/|\mathcal{G}_\tau|$, and $1/|\mathcal{B}|$ weight rollouts equally within each group and groups equally within each step. 

\begin{figure}[t]
  \centering
  \includegraphics[width=\textwidth]{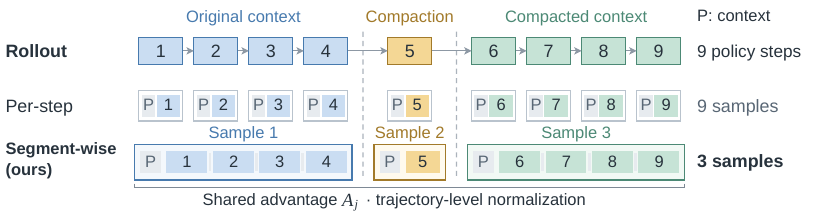}
  \caption{Segment-wise training across compaction. Colors denote maximal prefix-consistent segments, and step 5 performs compaction. Gray blocks indicate masked input context. Segment samples share the trajectory advantage $A_j$ and trajectory-level loss normalization. A 9-step rollout with one compaction event at step~5 yields 9 independent samples under per-step training but only 3 segment samples under our approach.}
  \label{fig:compaction_alignment}
\end{figure}

\subsection{Segment-wise Training across Compaction}
\label{appendix:segment_training}

During inference, the OpenCode harness applies auto context compaction when the accumulated context approaches the window limit: earlier turns are summarized into a compressed representation, and the agent continues with a fresh prefix. This enables arbitrarily long rollouts within a fixed context budget but introduces a structural challenge for training: the trajectory is no longer a single causal sequence.

Consider rollout $j$ of task $\tau$, producing $T_j$ agent steps. Let $o_t$ denote the tokens generated by the policy $\pi_\theta$ at step $t$. Between consecutive steps, the harness appends the environment's response (e.g., tool outputs) to the context, while when the context exceeds a threshold, compaction replaces it with a summary. We define a \textit{segment} as a maximal contiguous subsequence of steps whose contexts form a causal chain, \ie, no compaction occurs between them. Compaction events partition rollout $j$ into $M_j$ segments $S_1, \ldots, S_{M_j}$, each with a shared base prefix $p_m$, as shown in Figure~\ref{fig:compaction_alignment}.

A standard per-step decomposition treats each step as an independent sample with its own full harness context $x_t$ that is common in GUI agent training:
\begin{equation}
  \mathcal{L}_j^{\text{step}}(\theta) = -\frac{A_j}{L_j} \sum_{t=1}^{T_j} \log \pi_\theta(o_t \mid x_t),
  \label{eq:per_step_loss}
\end{equation}
where $L_j = \sum_{t=1}^{T_j} |o_t|$ is the total generated token count across the trajectory and $A_j$ is the trajectory advantage (Eq.~\ref{eq:advantage}). This requires $T_j$ forward passes with largely overlapping prefixes and discards the autoregressive dependency between consecutive steps sharing the same context window.
Our segment-wise formulation groups steps by their shared prefix and preserves the within-segment causal structure:
\begin{equation}
  \mathcal{L}_j^{\text{seg}}(\theta) = -\frac{A_j}{L_j} \sum_{m=1}^{M_j} \sum_{t \in S_m} \log \pi_\theta\!\left(o_t \,\middle|\, p_m,\, c_{m}^{<t}\right),
  \label{eq:segment_loss}
\end{equation}
where $c_{m}^{<t}$ denotes the in-segment history preceding step $t$: the interleaved sequence of prior policy outputs and environment responses from all earlier steps within segment $S_m$.
This design offers three benefits: 
(1) \textbf{Efficiency.} The number of forward passes reduces from $T_j$ to $M_j$ ($M_j \ll T_j$ in practice), avoiding redundant prefix encoding.
(2) \textbf{Training--inference alignment.} Within each segment, steps condition on the model's own prior outputs, matching the autoregressive generation process at inference time.
(3) \textbf{Balanced normalization.} Trajectory-level normalization by $L_j$ ensures that rollouts with frequent compactions (many short segments) receive the same total gradient weight as those with infrequent compactions (few long segments).

\section{Policy Behavior Study: Details}
\label{appendix:behavior}
\begin{table}[t]
  \caption{Key behavior statistics. OOD: 8 benchmarks pooled; net tasks: fail-to-pass minus pass-to-fail tasks relative to Base (QwenClawBench excluded). \ours-val: 33 tasks $\times$ 8 rollouts.}
  \label{tab:behavior_key}
  \small
  \setlength{\tabcolsep}{4pt}
  \begin{center}
  \begin{tabular}{@{}lrrrrr@{}}
    \toprule
    & Base & GLM-5.2 & +RFT & +SFT & +SFT+RL \\
    \midrule
    \multicolumn{6}{@{}l}{\emph{OOD benchmarks}} \\
    \midrule
    \quad Pass rate on common tasks (\%) & 38.7 & 54.5 & 44.6 & 46.2 & 48.4 \\
    \quad Net tasks gained over Base & -- & +207 & +73 & +93 & +129 \\
    \quad Interfaces used per episode & 2.40 & 2.65 & 2.36 & 2.61 & 2.68 \\
    \quad Interface switch rate (\%) & 18.4 & 26.3 & 15.8 & 24.7 & 25.5 \\
    \quad GUI tools (\% of tool calls) & 16.3 & 13.6 & 17.2 & 4.3 & 1.6 \\
    \quad Turn with $\geq$8K output tokens (\%) & 12.8 & 7.4 & 11.7 & 24.3 & 9.1 \\
    \midrule
    \multicolumn{6}{@{}l}{\emph{\ours-val (in-domain)}} \\
    \midrule
    \quad \passatk{1} (\%) & 5.7 & 37.9 & 10.3 & 37.3 & 36.0 \\
    \quad Interface switch rate (\%) & 26.1 & 34.1 & 25.8 & 37.0 & 35.7 \\
    \bottomrule
  \end{tabular}
  \end{center}
\end{table}

\myparagraph{Cross-interface interaction is inherited from the teacher.} After SFT, the agent uses more interfaces per episode (2.40 for the base model, 2.61 after SFT, 2.68 after RL) and rarely drives the GUI actions (16.3\%, 4.3\%, and 1.6\% of tool calls), whereas RFT keeps the base model's habits (2.36 and 17.2\%; Table~\ref{tab:behavior_key}). The same holds in-domain: on \ours-val, the switch rate is 26.1\% for the base model and 25.8\% for RFT, but 37.0\% for SFT, 35.7\% for \ours, and 34.1\% for GLM-5.2.

\myparagraph{The benefit of RL lies in transfer, not in-domain fit.} On \ours-val, SFT already matches the teacher (\passatk{1} 37.3\% vs.\ 37.9\%) and RL does not improve on it (36.0\%), yet RL raises the net OOD gain over the base model from $+93$ to $+129$ tasks ($+73$ for RFT). RL also removes the long turns that SFT produces but the teacher does not: a turn with at least 8K output tokens occurs in 24.3\% of SFT episodes, 9.1\% after RL, and 7.4\% for GLM-5.2.

\myparagraph{Where the step-budget gain arises.} Figure~\ref{fig:behavior_step_budget} pools CUG, OSV, and WA, where RL adds most over SFT (72 vs.\ 30 net tasks over the base model). On each of them, \ours gains more than SFT between 40 and 120 steps (CUG: 18.0 vs.\ 7.5; OSV: 16.2 vs.\ 8.1; WA: 16.4 vs.\ 12.4 points). Counting timeouts as unsolved is conservative for \ours: 5.5\% of its episodes on these benchmarks time out before step 120.

\myparagraph{What does not improve.} The longest tasks remain out of reach: the hardest quartile of Figure~\ref{fig:behavior_difficulty} consists mostly of OS-2 and ALE tasks, of which the base model and its four variants solve 4--7 and the strongest reference, GPT-6-Astra, 18. In future work, given sufficient compute resources, we plan to investigate whether larger models can better exploit our synthesized data and achieve stronger performance on these tasks after training.

\section{Case Study}
\label{appendix:case_study}

This section presents the persona and story of one synthesized world, the skills used by the generator in world and task synthesis, and an accepted task from the corpus. Table~\ref{tab:task_trajectory} traces an agent trajectory on this task, and Table~\ref{tab:verification} shows the synthesized verifier.

\myparagraph{Persona and Story.}
Table~\ref{tab:persona_case} shows the persona of one synthesized world. The persona profile specifies a freelance audio-visual technician, six named contacts, the websites and MCP services he uses, and three story arcs, one of which requires choosing products on Amazon by their images. The day planner expands these arcs into day plans, which are materialized into 17 world states. Each plan continues the open threads of the previous world state, such as the return request filed on day 0 and closed within its deadline on day 4, and the sports-bar commissioning that is prepared from day 1 and carried out on day 16. The applications used by the persona also grow over the days, as email, Slack, and the library service are added on days 1, 4, and 7. Four tasks are accepted on the world states of days 5, 7, and 9.

\begin{table}[t]
  \caption{Persona and story, condensed from the persona file and the day plans. The task of Table~\ref{tab:task_case_study} is synthesized on the world state of day 7.}
  \label{tab:persona_case}
  \centering
  \begin{adjustbox}{max width=\linewidth}
  \begin{tabularx}{\linewidth}{@{}l>{\raggedright\arraybackslash}X@{}}
    \toprule
    \multicolumn{2}{@{}l}{\textit{Persona}} \\
    Profile & Lars Petersen, 41, freelance audio-visual technician and sole proprietor of an AV business in Minneapolis. He installs and maintains display walls, audio, and projection for a sports bar, a church, and a community theatre. \\
    Contacts & Marco Delgado (sports-bar owner), Pastor Tom Hauge (church client), Dee Nguyen (distributor account manager), Brendan Olsrud (install assistant), Ingrid and S{\o}ren Petersen (wife and son). \\
    Applications & Websites: ESPN, Amazon, Google Maps. MCP services: Support Tickets, Notifications, Direct Messages, Calendar. \\
    Story arcs & (1) Choose four 65-inch displays and rack surge protectors on Amazon by their product images for the sports-bar display wall. (2) Schedule commissioning around the televised games of April 11 and 12 using ESPN. (3) Plan the driving route between suppliers and client sites on Google Maps. \\
    \midrule
    \multicolumn{2}{@{}l}{\textit{Day plans (17 world states, 2024-03-25 to 2024-04-10)}} \\
    Day 0 & Initial state. The return request for a failed surge protector at the church is filed with a deadline on day 4. \\
    Day 1 & Mount brackets and pull cables at the sports bar. After the owner approves by direct message, order the displays, streaming sticks, and surge protectors on Amazon. \\
    Days 2--4 & Build the rack, receive and mount the displays, install the replacement surge protector at the church, and close the return request within its deadline. \\
    Days 5--6 & Coach the son's hockey game, check the broadcast schedule on ESPN, and run the Sunday service check at the church. \\
    Day 7 & Pick up a replacement projector from the distributor and prepare the sports-bar commissioning with the assistant. \\
    Days 8--15 & Install the replacement projector at the theatre and close its support ticket (day 11), run the weekly church check (day 13), and prepare the commissioning (days 14--15). \\
    Day 16 & Commission the sports-bar display wall before the televised games of April 11 and 12. \\
    \midrule
    \multicolumn{2}{@{}l}{\textit{State growth from day 0 to day 16}} \\
    MCP services & From 4 to 7 services, with email added on day 1, Slack on day 4, and Library on day 7. MCP database rows grow from 99 to 344. \\
    Websites & Accounts on ESPN, Amazon, Google Maps, and Google Search on day 0, AllRecipes from day 1, and WolframAlpha from day 3. Amazon orders grow from 0 to 2, ESPN favorites from 3 to 10, and Google Maps timeline entries from 2 to 28. \\
    \midrule
    \multicolumn{2}{@{}l}{\textit{Accepted tasks (4)}} \\
    Day 5 & Rank the open evenings of the coming week, save a dinner recipe chosen by its photo on AllRecipes, add a calendar block, and message his wife. \\
    Day 7 & Two tasks: reorder the rack surge protector that is at its reorder level (\texttt{d7-h1}), and the task of Table~\ref{tab:task_case_study} (\texttt{d7-h2}). \\
    Day 9 & Score in-stock surge protectors from the return log, post a review on the product identified by its image, and reply to the distributor. \\
    \bottomrule
  \end{tabularx}
  \end{adjustbox}
\end{table}

\myparagraph{Synthesis Skills.}
In world synthesis, the generator matches the activities of each day plan against a shared library of event creation skills (see Sec.~\ref{sec:method:world}). The library contains 83 skills in total, and Table~\ref{tab:world_skills} lists 20 of them. Each of these skill files consists of a description, a parameter table, one scenario per involved application with the database records or local files to produce, consistency checks against the existing state, preconditions, and possible follow-up events. 
Table~\ref{tab:world_skill_case} further shows the content of \texttt{book\_flight} skill. Its five scenarios write a booking on the Google Flights website, a confirmation email, a direct-message exchange, calendar events, and a local notes file, and the same confirmation number appears in all five applications. Its consistency checks read the persona state and the existing records before insertion, so that message order, calendar entries, and existing files remain consistent with earlier days.

In task synthesis, the generator's evolving skills are short notes that the generator model derives from synthesis events (Sec.~\ref{sec:method:task}). An event is recorded when a candidate is returned for deepening after being judged too easy, for easing after no rollout succeeds, or for re-authoring after a successful rollout skips a required interface, and when the candidate reaches its final verdict. 
For each event, the generator reads the gate verdict, the rollout statistics, the feedback, and the existing notes, and returns at most two updates or no update. 
Notes are limited to 2,000 characters and 50 files to avoid overfitting to a specific scenario or task type.
The generator is instructed to edit existing notes before creating new ones. 
Across the task synthesis, we have 802 revisions of the 40 final skills. 
Table~\ref{tab:task_skills} shows 20 of these skills by task difficulty, interface coverage, write persistence, and verifier reliability. 
Table~\ref{tab:task_skill_case} further traces \texttt{hard-multi-interface-recipe} that is triggered by 40 tasks. This skill evolves from a simple recipe task to a complex multi-interface task, which can benefit the generator in future task synthesis. 

\begin{table}[t]
  \caption{World synthesis skill examples and the applications involved in each skill. Every listed skill also writes local files through the CLI. DM means Direct Messages.}
  \label{tab:world_skills}
  \centering
  \begin{tabularx}{\linewidth}{@{}ll>{\raggedright\arraybackslash}X@{}}
    \toprule
    Skill & GUI & MCP \\
    \midrule
    \multicolumn{3}{@{}l}{\textit{GUI, MCP, and CLI}} \\
    \texttt{japan\_trip\_research} & Google Flights, Booking & Notifications, DM, Calendar \\
    \texttt{book\_flight} & Google Flights & Email, DM, Calendar \\
    \texttt{research\_hotels} & Booking & DM, Notifications \\
    \texttt{japan\_hotel\_booking} & Booking & DM, Email \\
    \texttt{amazon\_purchase} & Amazon & Email, Calendar \\
    \texttt{coursera\_study\_session} & Coursera & Calendar, DM \\
    \texttt{attend\_sports\_event} & ESPN & Calendar, DM \\
    \texttt{meal\_prep} & AllRecipes & DM \\
    \midrule
    \multicolumn{3}{@{}l}{\textit{MCP and CLI}} \\
    \texttt{attend\_lab\_session} & -- & Calendar, Slack, Canvas LMS \\
    \texttt{prepare\_lab\_submission} & -- & Canvas LMS, Slack \\
    \texttt{debug\_ci\_failure} & -- & Slack, Email, Support Tickets \\
    \texttt{on\_call\_incident} & -- & Email, Slack, DM \\
    \texttt{code\_review} & -- & Slack, Email \\
    \texttt{configure\_feature\_flag} & -- & Slack, Calendar, Email \\
    \texttt{respond\_to\_vendor\_email} & -- & Email, Support Tickets \\
    \texttt{awit\_event\_registration} & -- & Calendar, Email, Slack \\
    \texttt{payroll\_approval} & -- & Email, Calendar, DM, Notifications \\
    \texttt{schedule\_equipment\_service} & -- & Calendar, Email, DM \\
    \texttt{complete\_permit\_renewal} & -- & Email, Calendar, DM \\
    \texttt{reply\_to\_customer\_review} & -- & Email, Calendar \\
    \bottomrule
  \end{tabularx}
\end{table}

\begin{table}[t]
  \caption{World synthesis skill \texttt{book\_flight}, condensed from the skill file. Each scenario specifies the state changes in one application. }
  \label{tab:world_skill_case}
  \centering
  \begin{tabularx}{\linewidth}{@{}l>{\raggedright\arraybackslash}X@{}}
    \toprule
    Section & Content \\
    \midrule
    Description & The user books a flight on Google Flights, receives the airline confirmation, shares it with the travel companion by direct message, and records the trip in the calendar and a notes file. \\
    Parameters & 21 parameters, \eg, route AUS--NRT, carrier JAL, fare 1,089 USD, seat 34A, confirmation number, message thread 2, and notes path. \\
    \midrule
    GUI: Google Flights & Read the app schema, insert a booking row with status \texttt{confirmed}, and verify it by the confirmation number. \\
    MCP: Email & Insert the airline confirmation thread and message into the personal inbox with the Inbox, Updates, and Japan-Trip labels. \\
    MCP: Direct Messages & Read the latest messages in thread 2, insert the user's message and the companion's reply, and update the thread timestamp. \\
    MCP: Calendar & Check for existing trip events, then insert events for the outbound flight, the trip, and the return flight into calendar 1. \\
    CLI: Files & Check the notes directory, then create a notes file of 60--100 lines with flight details, trip plan, to-do list, and budget. \\
    \midrule
    Consistency checks & Read identities from the persona state file, insert messages in chronological order without duplicates, update existing calendar placeholders, look up the label identifier, keep existing notes files, and place the companion's reply 30--90 minutes after the user's message. \\
    Preconditions & The trip storyline is active, and the Google Flights account (created if missing), message thread 2, email label 12, and calendar 1 exist. \\
    Follow-ups & Browse and book hotels, share an itinerary document, check email, block out-of-office time, and order travel gear. \\
    \bottomrule
  \end{tabularx}
\end{table}

\begin{table}[t]
  \caption{Evolving task synthesis skills, grouped by the task property they address. Lessons are paraphrased from the skill text. Revisions: number of revisions made during the synthesis.}
  \label{tab:task_skills}
  \centering
  \begin{tabularx}{\linewidth}{@{}>{\raggedright\arraybackslash}Xr@{}}
    \toprule
    Skill and lesson & Revisions \\
    \midrule
    \multicolumn{2}{@{}l}{\textit{Task difficulty}} \\
    \midrule
    \texttt{image-pick-decisive-cue}: Naming the dish, the visual cue, or a category with a color reduces the image choice to text search. & 88 \\
    \texttt{hard-multi-interface-recipe}: Two strong steps make a task hard, and a CLI computation counts only if later steps depend on it. & 43 \\
    \texttt{image-only-sourcing-not-leak-protection}: Criteria that match search filters or indexed text identify the target without image inspection. & 35 \\
    \texttt{criteria-specified-parse-is-lookup}: Criteria that uniquely identify the target turn parsing into a lookup. & 13 \\
    \texttt{shortcut-resistance-not-gate-count}: A multi-step CLI computation counts as one step, and naming its output removes it. & 10 \\
    \texttt{answer-key-saturation-strip-before-volume}: Naming every resolved value, or all candidates except one, removes the need for inspection. & 7 \\
    \texttt{difficulty-calibration-intrinsic}: Difficulty is calibrated on unhinted rollouts and raised through more required work. & 5 \\
    \midrule
    \multicolumn{2}{@{}l}{\textit{Interface coverage}} \\
    \midrule
    \texttt{multi-interface-indispensable}: An MCP message that restates fabricable values can be skipped unless it depends on live values absent from the CLI steps. & 79 \\
    \texttt{web-state-restated-post-bypass}: A message that restates website values can be fabricated without visiting the website unless a visual decision requires the visit. & 55 \\
    \texttt{interface-flow-answer-key-leak}: An MCP service that only provides context and receives a restating message can be skipped. & 39 \\
    \texttt{answer-key-leak-cascades-interface-bypass}: A CLI step that names the target lets agents skip the website and MCP steps. & 27 \\
    \texttt{chained-decisive-step-handoff}: Verifiers check the live website row and message, since checking a CLI copy lets agents skip both. & 24 \\
    \texttt{indispensability-not-hardness}: Requiring an interface keeps it in the trajectory without making the task hard. & 14 \\
    \midrule
    \multicolumn{2}{@{}l}{\textit{Write persistence}} \\
    \midrule
    \texttt{browser-write-surface-persistent}: A website write that shows success without a new database row makes the task unsolvable. & 47 \\
    \texttt{answer-keyed-zero-is-infrastructure}: Zero completions at both hint tiers are first checked for a non-persistent write. & 31 \\
    \texttt{web-save-preexisting-list-false-positive}: Saving to a list that already exists lets a stale row pass the check. & 16 \\
    \midrule
    \multicolumn{2}{@{}l}{\textit{Verifier reliability}} \\
    \midrule
    \texttt{verifier-communication-mention-tokens}: Message checks require the concrete values requested by the task, such as amounts and dates. & 14 \\
    \texttt{verifier-dm-thread-lookup}: Message threads are resolved by topic, since one participant can appear in several threads. & 3 \\
    \texttt{verifier-evidence-machine-checkable}: Checks target new or modified state and identify items by database keys. & 2 \\
    \texttt{verifier-script-crash-is-code-bug}: A verifier crash on the initial or reference-solution state is fixed in the script. & 2 \\
    \bottomrule
  \end{tabularx}
\end{table}

\begin{table}[t]
  \caption{Revisions of the evolving task synthesis skill \texttt{hard-multi-interface-recipe}, listed in order with the gate outcome that triggered each revision. Some intermediate revisions are omitted for brevity. The skill text is paraphrased for human readability.}
  \label{tab:task_skill_case}
  \centering
  \begin{tabularx}{\linewidth}{@{}ll>{\raggedright\arraybackslash}X@{}}
    \toprule
    Revision & Trigger & Change \\
    \midrule
    1 & Accepted as hard & Place the decisive difficulty in a computational CLI step, tie each value passed between interfaces to an upstream record, and make every interface change the world state. \\
    2 & Accepted as hard & A visual choice identified by at least two cues unique to the target is an alternative source of difficulty. \\
    3 & Returned as too easy & A parse-and-rank CLI step combined with a visual choice that reduces to reading labels remains too easy. \\
    6 & Accepted as hard & With a single visual choice, the computational step must integrate values from several interfaces in a flow that returns to an earlier interface. \\
    11 & Accepted as hard & A non-CLI read that matches the ranked result against another person's choice is a third source of difficulty. \\
    14 & Returned as too easy & Naming the output of any such step in the task removes its difficulty, so the task states criteria and leaves resolved values unnamed. \\
    19 & Accepted as hard & Two strong steps, a multi-source computation and a visual choice by art style without a text equivalent, make a task hard. \\
    32 & Returned as too easy & A computational CLI step counts only if a later step depends on its result, and restating the result in a later file does not count. \\
    43 & Returned as too easy & A value computed for a message counts only if the task gives neither the formula nor the result. \\
    \midrule
    Final & After revision 43 & A task is hard when it contains two strong steps, each being a computational CLI step, a visual choice unique to the target, a non-CLI read, or an inference. A computational step counts only if a later step depends on its result. A computation fed by live MCP or website reads yields a shortlist that agents cannot produce without those interfaces, and a visual decision within the shortlist is the second strong step. A final message that carries values from at least two independent sources requires all three interfaces. A non-persistent website save lets agents complete the task with the CLI alone. \\
    \bottomrule
  \end{tabularx}
\end{table}

\myparagraph{Agent Trajectory.}
Table~\ref{tab:task_trajectory} shows the task and six of the 65 tool calls in a rollout that completes it. The task does not name any interface, but its inputs and outputs are distributed across interfaces: the invoice is a PDF file in the local workspace, the purchase requires signing in to the Amazon website supported by WebHarbor~\citep{webharbor2026}, and the recipient is identified by a direct-message thread. The agent first resolves the recipient through the direct-message service (step~2) and extracts the prior order from the invoice with a Python script (step~9). It then searches the storefront, takes screenshots of candidate product pages, and queries the vision tool to check the product attributes specified in the request (step~45), as the listing text does not state them. After adding the selected speaker to the cart (step~59), the agent sends the product name and cart price to the recipient (step~62) and writes the selection back into the shortlist file (step~64). These two steps use information obtained in the browser and can therefore only be completed after the GUI sub-task.

\myparagraph{Verifier.}
Table~\ref{tab:verification} shows the synthesized verifier for this task. It consists of three scripts, one for each outcome requested by the task, and a rollout succeeds only if all scripts pass, \ie, $V=\prod_{i}v_i$ (Sec.~\ref{sec:method:task}). Each check compares the final state with the initial state through primitives shared by all synthesized verifiers, such as \texttt{web\_new\_rows}, which returns only the rows added during the rollout. 
A state that already satisfies the request in $W_0$ therefore does not pass the check, which is the property tested by the initial unsatisfaction gate. The checks identify their targets by natural keys, namely the product slug and the participant name of the direct-message thread. The last block compares two rollouts that add different speakers to the cart. Only the first rollout adds a row for the target product, so \texttt{check\_cart.py} passes for this rollout and fails for the second one, whose direct-message check also fails ($V=0$). 

\definecolor{xaicli}{HTML}{2F5C8F}
\definecolor{xaimcp}{HTML}{1A7F37}
\definecolor{xaigui}{HTML}{C2620A}
\definecolor{xaiclibg}{HTML}{E8EFF8}
\definecolor{xaimcpbg}{HTML}{E6F3EA}
\definecolor{xaiguibg}{HTML}{FBEEE1}
\definecolor{codebg}{HTML}{F2F3F5}
\definecolor{elidegrey}{HTML}{8A8F98}

\providecommand{\xcli}[1]{\textcolor{xaicli}{#1}}
\providecommand{\xmcp}[1]{\textcolor{xaimcp}{#1}}
\providecommand{\xgui}[1]{\textcolor{xaigui}{#1}}
\providecommand{\xtag}[2]{%
  \colorbox{#1}{\makebox[1.9em][c]{\textcolor{white}{\sffamily\bfseries\tiny #2}}}}
\providecommand{\tCLI}{\xtag{xaicli}{CLI}}
\providecommand{\tMCP}{\xtag{xaimcp}{MCP}}
\providecommand{\tGUI}{\xtag{xaigui}{GUI}}
\providecommand{\obs}[1]{%
  \colorbox{codebg}{\parbox[t]{0.425\textwidth}{\ttfamily\tiny\baselineskip=6.2pt #1\strut}}}
\providecommand{\elide}[1]{%
  \midrule \multicolumn{3}{@{}c@{}}{\textit{#1}} \\ \midrule}

\begin{table}[t]
  \caption{Case study of agent trajectory on XAI task. Color marks the part of the request that only a \xcli{shell}, only the \xgui{browser}, or only an \xmcp{MCP app} can proceed. Six of the trajectory's total 65 tool calls are shown with the observation the agent received and the action it took. }
  \label{tab:task_trajectory}
  \centering
  \scriptsize
  \setlength{\tabcolsep}{4pt}
  \renewcommand{\arraystretch}{1.2}
  \begin{tabularx}{\textwidth}{@{}>{\raggedright\arraybackslash}p{0.078\textwidth}>{\raggedright\arraybackslash}p{0.445\textwidth}>{\raggedright\arraybackslash}X@{}}
    \toprule
    \multicolumn{3}{@{}l@{}}{\textbf{User request}\;\;\small\textnormal{(\texttt{task.md})}} \\
    \midrule
    \multicolumn{3}{@{}p{\textwidth}@{}}{\itshape
      Wren (my climbing partner) and I share a ``wren -- climbing'' direct-message thread, and I want
      to send Wren a portable Bluetooth speaker as a gift. I just received a JBL Clip~4 from a recent
      Amazon order (the invoice PDF is at \texttt{Downloads/amazon\_invoice\_2026-04-02.pdf}), so I
      need to avoid buying the same unit again and stay around what that one cost.
      \par\smallskip
      \xcli{Write and run a script that parses that invoice PDF to pull the prior unit, its order
      number, and the delivery date, then create a new file
      \texttt{./gift\_candidate\_shortlist.csv} \ldots\ use its current Amazon price as the
      reference \ldots}
      \par\smallskip
      \xgui{Sign in to Amazon with the demo account shown on the login page and add the top-ranked,
      visually matching option to the cart --- the option whose visible design is a cylindrical body
      with a black mesh grille over solid smooth ends.}
      \par\smallskip
      \xcli{\ldots\ fold the selection back into the shortlist.}
      \xmcp{Then send Wren a direct message in our ``wren -- climbing'' thread confirming the gift is
      added to the cart, including the selected speaker's name and cart price \ldots}\strut} \\
    \midrule
    \textbf{Step} & \textbf{Observation} & \textbf{Action} \\
    \midrule
    \tMCP\;2 &
      \obs{\{"items": [\{"id": 2, "title": "wren - climbing",\\
      \hspace*{1em}"participants": [\{"display\_name": "Taylor"\},\\
      \hspace*{3.4em}\{"display\_name": "Wren Osei"\}]\}, \ldots]\}} &
      \texttt{sms\_list\_threads(search="Wren")}
      \newline\textit{The recipient is named only in prose, so the agent has to resolve the thread and
      the person before it can send anything.} \\
    \elide{6 tool calls omitted --- installs a PDF library, opens the storefront}
    \tCLI\;9 &
      \obs{Amazon Order Invoice - 2026-04-02\\
      Order \#114-20260331-042280 - JBL Clip 4\\
      Portable Bluetooth Speaker - delivered.} &
      \texttt{bash}: writes and runs a \texttt{pypdf} script over the invoice
      \newline\textit{Fixes the unit that must not be bought again and the \$49.95 price band the
      rest of the episode is held to.} \\
    \elide{35 tool calls omitted --- signs in, searches, vision-checks six candidate listings}
    \tGUI\;45 &
      \raisebox{\dimexpr\ht\strutbox-\height\relax}{\includegraphics[width=0.437\textwidth]{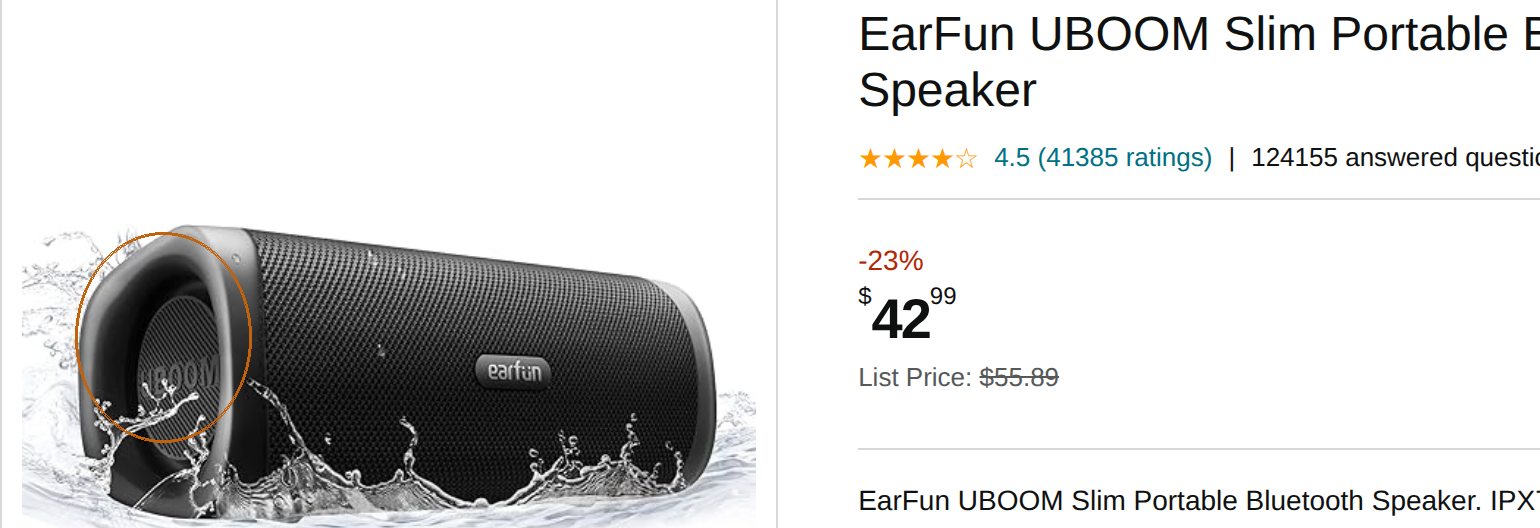}} &
      \texttt{take\_screenshot} $\rightarrow$ \texttt{describe\_image(} \texttt{"Is the body
      cylindrical? Is there a black mesh grille over solid smooth plastic ends?")}
      \newline\textit{Reply: ``the body is cylindrical\ldots the two end caps are solid, smooth black
      plastic''. The listing text says none of this.} \\
    \elide{13 tool calls omitted --- locates the buy box}
    \tGUI\;59 &
      \raisebox{\dimexpr\ht\strutbox-\height\relax}{\includegraphics[width=0.437\textwidth]{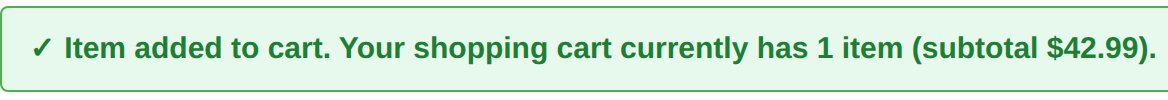}} &
      \texttt{click("Add to Cart")}
      \newline\textit{The one step in the episode that changes storefront state, and the one the
      verifier reads back.} \\
    \elide{2 tool calls omitted}
    \tMCP\;62 &
      \obs{\{"message\_id": 43, "thread\_id": 2,\\
      \hspace*{0.6em}"body": "\ldots EarFun UBOOM Slim \ldots Cart\\
      \hspace*{2.6em}price \$42.99 \ldots added to cart",\\
      \hspace*{0.6em}"sent\_at": "2026-04-07T10:13:08"\}} &
      \texttt{sms\_send(thread\_id=2, body=\ldots)}
      \newline\textit{Closes on the interface the episode opened on, carrying a fact that only the browser
      actions could establish.} \\
    \elide{1 tool call omitted}
    \tCLI\;64 &
      \obs{rank,candidate\_label,max\_price\_usd,\ldots\\
      1,"EarFun UBOOM Slim\ldots",49.95,\\
      \hspace*{0.6em}"Cylindrical (black mesh grille over\\
      \hspace*{0.6em}solid smooth ends)",false,"CHOSEN -\\
      \hspace*{0.6em}\$42.99 (added to Amazon cart)"} &
      \texttt{write("gift\_candidate\_shortlist.csv")}
      \newline\textit{Folds the browser outcome back into the shell artifact, so the CSV can only be
      completed after the cart update in the browser.} \\
    \bottomrule
  \end{tabularx}
\end{table}

\lstdefinestyle{verifierstyle}{
  language=Python,
  basicstyle=\ttfamily\scriptsize,
  keywordstyle=\color{xaicli}\bfseries,
  stringstyle=\color{xaigui},
  commentstyle=\color{elidegrey}\itshape,
  backgroundcolor=\color{codebg},
  frame=single, framerule=0pt, framesep=5pt, rulecolor=\color{codebg},
  xleftmargin=5pt, xrightmargin=5pt, framexleftmargin=0pt,
  breaklines=true, breakindent=12pt, columns=fullflexible, keepspaces=true,
  showstringspaces=false, aboveskip=3pt, belowskip=3pt, numbers=none,
}

\begin{table}[t]
  \caption{Case study on verification. The verifier uses a checklist of multiple scripts  to check whether the agent's rollout produced the expected changes to the world state.}
  \label{tab:verification}
  \centering
  \scriptsize
  \setlength{\tabcolsep}{4pt}
  \renewcommand{\arraystretch}{1.2}

  \begin{tabularx}{\textwidth}{@{}l l >{\raggedright\arraybackslash}X@{}}
    \toprule
    Criterion & Surface & Evidence required in the final world \\
    \midrule
    \texttt{check\_shortlist.py} & \tCLI &
      a \emph{new} workspace file \texttt{gift\_candidate\_shortlist.csv} carrying the order number and
      delivery date read out of the PDF, the price reference, and the folded selection \\
    \texttt{check\_cart.py} & \tGUI &
      a \emph{new} \texttt{cart\_items} row whose product resolves from the target's slug \\
    \texttt{check\_dm.py} & \tMCP &
      a \emph{new} message in the thread whose participant name contains ``Wren'', mentioning the
      speaker, the cart price and the cart \\
    \midrule
    \multicolumn{3}{@{}l@{}}{\textbf{Browser Check}} \\
    \bottomrule
  \end{tabularx}
  \vspace{2pt}
\begin{lstlisting}[style=verifierstyle]
EARFUN_SLUG = "earfun-uboom-slim-portable-bluetooth-speaker"

r.check("earfun_cart_row",
        lambda: w.web_new_rows("amazon", "cart_items",
            where="product_id = (SELECT id FROM products WHERE slug = ?)",
            params=(EARFUN_SLUG,)))
\end{lstlisting}

  \vspace{1pt}
  \begin{tabularx}{\textwidth}{@{}X@{}}
    \toprule
    \textbf{The primitive it is built on}\;\;{\small (shared by every synthesized verifier)} \\
    \bottomrule
  \end{tabularx}
  \vspace{2pt}
\begin{lstlisting}[style=verifierstyle]
def web_new_rows(self, site, table, where="1=1", params=()):
    base = {r["_pk"] for r in self.baseline.web(site, f"SELECT {pk} AS _pk FROM {table}")}
    rows = self.after.web(site, f"SELECT *, {pk} AS _pk FROM {table} WHERE {where}", params)
    return [r for r in rows if r["_pk"] not in base]   # after minus baseline
\end{lstlisting}

  \vspace{2pt}
  \begin{tabularx}{\textwidth}{@{}>{\columncolor{white}[0pt][\tabcolsep]\raggedright\arraybackslash}p{0.30\textwidth} >{\raggedright\arraybackslash}X >{\columncolor{white}[\tabcolsep][0pt]}c@{}}
    \toprule
    \multicolumn{3}{@{}l@{}}{\textbf{What the check separates}} \\
    \midrule
    \textbf{Rollout} & \textbf{\texttt{amazon.cart\_items}, after $-$ before} & \textbf{$V$} \\
    \midrule
    \rowcolor{xaimcpbg}
    carts the speaker the product image matches &
      \texttt{\{user\_id:\,1, product\_id:\,235, quantity:\,1\}} \;\xmcp{$\checkmark$} & 1 \\
    \midrule
    \rowcolor{xaiguibg}
    carts a speaker that is plausible from the listing text alone &
      \texttt{\{user\_id:\,1, product\_id:\,419, quantity:\,1\}} \;\xgui{$\times$} \;
      \textnormal{--- no row for the target slug, so the message that reports it fails too} & 0 \\
    \bottomrule
  \end{tabularx}
\end{table}

\end{document}